\documentclass[%
 aip,jcp, 
 amsmath,amssymb,
 reprint,%
]{revtex4-2}
\usepackage{graphicx}
\usepackage{dcolumn}
\usepackage{bm}

\usepackage[utf8]{inputenc}
\usepackage[T1]{fontenc}
\usepackage{mathptmx}
\usepackage{etoolbox}
\usepackage{enumitem}
\usepackage{subfigure}
\usepackage{hyperref}

\usepackage{xcolor}

\newcommand{\nb}{$\mathbf{N}_{b}$\ }

\makeatletter
\def\@email#1#2{%
 \endgroup
 \patchcmd{\titleblock@produce}
  {\frontmatter@RRAPformat}
  {\frontmatter@RRAPformat{\produce@RRAP{*#1\href{mailto:#2}{#2}}}\frontmatter@RRAPformat}
  {}{}
}%
\makeatother
\begin{document}

\preprint{AIP/123-QED}

\title{Active Learning of Molecular Data for Task-Specific Objectives}

\author{Kunal Ghosh}
\affiliation{Department of Applied Physics, Aalto University, P.O. Box 11000, FI-00076 Aalto, Finland}
 \affiliation{Department of Computer Science, Aalto University, P.O. Box 15400, FI-00076 Aalto, Finland}
 
\author{Milica Todorovi\'{c}}%
\affiliation{Department of Mechanical and Materials Engineering, University of Turku, FI-20014 Turku, Finland}

\author{Aki Vehtari}
 \affiliation{Department of Computer Science, Aalto University, P.O. Box 15400, FI-00076 Aalto, Finland}

\author{Patrick Rinke}
    \email{patrick.rinke@tum.de}
\affiliation{Department of Applied Physics, Aalto University, P.O. Box 11000, FI-00076 Aalto, Finland}
\affiliation{Physics Department, TUM School of Natural Sciences, Technical University of Munich, Garching, Germany}
\affiliation{Atomistic Modelling Center, Munich Data Science Institute, Technical University of Munich, Garching, Germany}

\date{\today}

\begin{abstract}
Active learning (AL) has shown promise for being a particularly data-efficient machine learning approach. Yet, its performance depends on the application and it is not clear when AL practitioners can expect computational savings. Here, we carry out a systematic AL performance assessment for three diverse molecular datasets and two common scientific tasks: compiling compact, informative datasets and targeted molecular searches. We implemented AL with Gaussian processes (GP) and used the many-body tensor as molecular representation. For the first task, we tested different data acquisition strategies, batch sizes and GP noise settings. AL was insensitive to the acquisition batch size and we observed the best AL performance for the acquisition strategy that combines uncertainty reduction with clustering to promote diversity. However, for optimal GP noise settings, AL did not outperform randomized selection of data points. Conversely, for targeted searches, AL outperformed random sampling and achieved data savings up to 64\%. 
Our analysis provides insight into this task-specific performance difference in terms of target distributions and data collection strategies. We established that the performance of AL depends on the relative distribution of the target molecules in comparison to the total dataset distribution, with the largest computational savings achieved when their overlap is minimal. 

\end{abstract}

\maketitle

\section{Introduction}
\noindent In recent years, applications of machine learning (ML) in material science have produced a plethora of new discoveries\cite{kulik_roadmap_2022,himanen_datadrivenmatSci_2019, bartok_machine_2017}. These discoveries rely on accurate property predictions by ML models, which usually require large amounts of training data\cite{domingos_cacm_2012, schutt_schnet_2017, ghosh_deep_2019}. Such large training datasets are costly to compile for supervised ML tasks\cite{westermayr_physically_2021,atz_geometric_2021,behler_constructing_2015,westermayr_high-throughput_2023}, because the property labels are obtained from expensive simulations or time-consuming experiments.

\begin{figure}[!htbp]
    \centering
    \includegraphics[width=\linewidth]{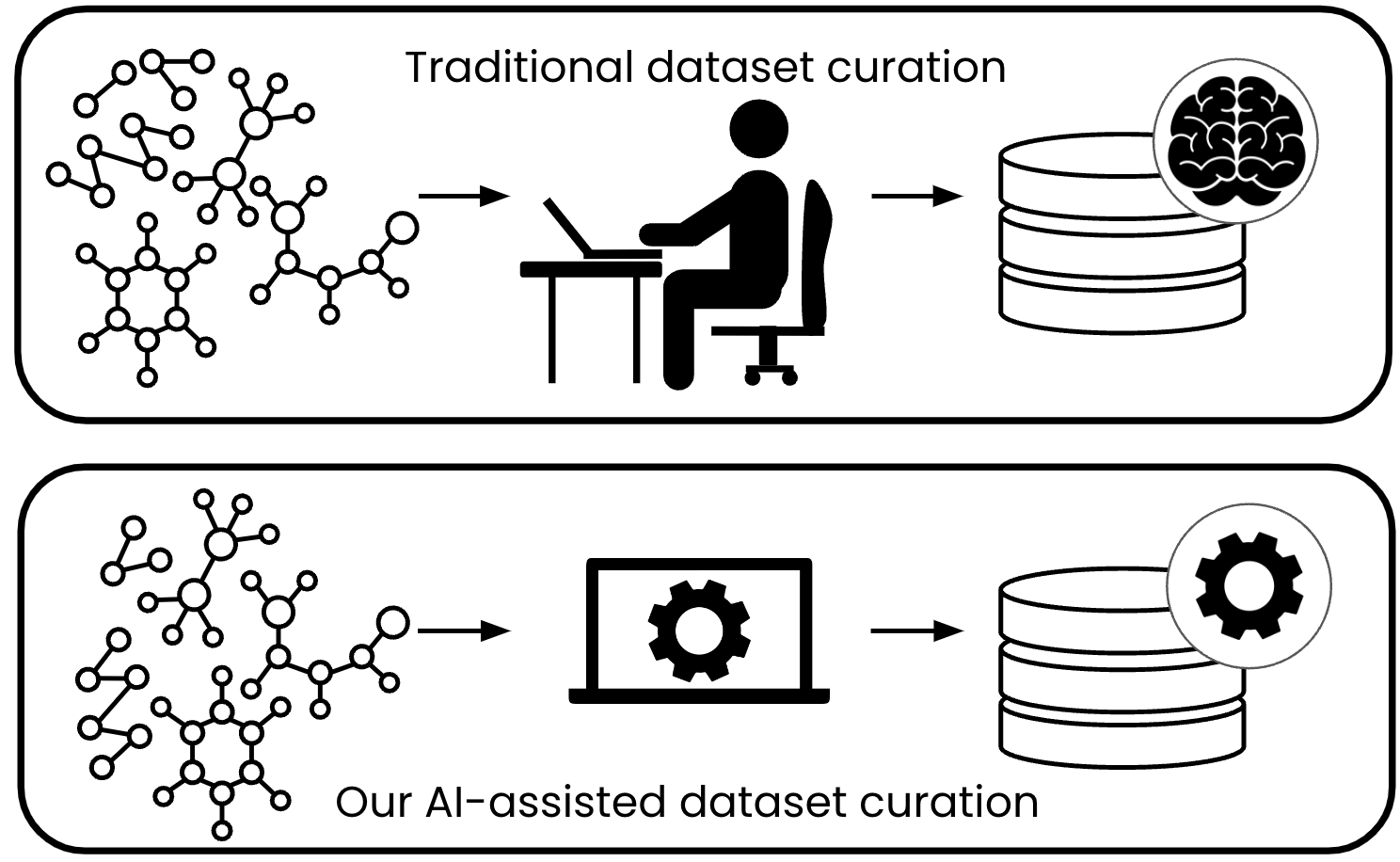}
    \caption{Traditionally, materials datasets are curated combinatorially, using human intuition. We propose an AI-assisted dataset curation scheme}
    \label{fig:fig1}
\end{figure}

Materials datasets are mostly compiled by human experts\cite{ramakrishnan_quantum_2014}, which can lead to bias and redundancy \cite{stuke_chemical_2019}. Instead of collecting larger training datasets to mitigate biases, we propose the use of AL to curate smaller datasets that are more informative. AL iteratively improves the performance of ML models by intelligently curating the model's training data via acquisition functions. The strategic data-driven curation reduces dataset redundancies in comparison to human curation. Moreover, different acquisition strategies can be exploited to produce datasets specifically designed for targeted ML tasks (see Figure \ref{fig:fig1}).

In materials science, AL has been deployed in development of novel algorithms\cite{fujiwara_jcim_2008, reker_multi-objective_2016, zhou_jointly_2022, hwang_uncertainty-based_2022} and applications to material discovery and property prediction tasks\cite{desai_rapid_2013, besnard_automated_2012, naik_active_2016, viet_johansson_using_2022, vandenhaute_machine_2023}. While AL holds considerable promise for smart data collection\cite{wen_2023, jose_2023, besel_2024},
just as often failure or no improvements are reported (some of these reports are anecdotal since failures are rarely published) \cite{zaverkin_exploration_2021, richards_2011, farquhar_2021}. 
To address the apparent discrepancy, we present a systematic study into the performance and benefits of AL for curation of molecular datasets. We compare the benefits of AL based on Gaussian process regression (GPR) in dataset curation for two different ML tasks: dataset pruning and inverse material design.

The objective of the first task is to generate a maximally compact and informative dataset (task 1). In this very common AL use case, one seeks to balance the redundancy and diversity of materials data. 
We propose batch acquisition strategies (AS) for AL and assess their performance for a variety of parameter choices and different materials datasets. We seek to identify the acquisition strategies which generate the smallest training sets and lead to the best model performance.

In the second dataset curation task, we address targeted materials property search (task 2). This objective emulates the ML use case of backward prediction: given a property target, we seek the best candidate materials with this feature. For the ML model to accurately predict the feature, the AL-curated dataset should balance data entries with and without this information. We use the AS approaches from task 1 to explore different routes towards dataset assembly for this task. Our objective here is to identify the approach that maximizes the predictive accuracy of the ML model on task 2.

This study employs three molecular datasets with different levels of complexity and redundancy \cite{stuke_chemical_2019}. We focus on learning the ionization potential, equivalent to the energy of the highest occupied molecular orbital (HOMO) computed by \textit{ab-initio} simulations. In task 2, we seek to identify molecules with a target property: HOMO values greater than $\varepsilon$. We make use of pre-labeled datasets to accelerate our study: starting from the large pool of possible molecular structures, we draw data points with AL and include labels into the ML model. This is analogous to a realistic AL use case, where researchers might start with a large pool of unlabeled materials structures, and perform computations to label them as needed\cite{besel_2024}. Smaller training datasets would require fewer calculations to obtain the material property labels for ML training. Therefore, the outcome of task 1 is to demonstrate whether computational savings can be achieved by generating maximally informative, compact datasets. The outcome of task 2 is to compile a list of materials that match a targeted property, and consequently identify optimal materials for a technological application. 

\section{Datasets}
\textbf{AA (44k amino acids and dipeptides):} The amino acid (AA) dataset\cite{ropo_first-principles_2016} contains 44 004 isolated and cation-coordinated conformers of 20 proteinogenic amino acids and their amino-methylated and acetylated (capped) dipeptides. The molecules reach up to 39 atoms in size and include the chemical elements H, C, N, O, and S as well as divalent cations (Ca2+, Ba2+, Sr2+, Cd2+, Pb2+, and Hg2+). The amino acid conformers encode different protonation states of the backbone and the side chains. Since all amino acids share a common backbone, the complexity of this dataset lies in differing side chains, dihedral angles and metal cations.  The AA dataset was generated by conformational sampling and all molecular structures and properties were calculated with density-functional theory (DFT) using the Perdew-Burke-Ernzerhof (PBE)\cite{perdew_generalized_1997} exchange-correlation functional with Tkatchenko-Scheffler van der Waals corrections (vdW)\cite{tkatchenko_accurate_2009} AA was used to benchmark several ML models \cite{bartok_machine_2017, de_comparing_2016, artrith_efficient_2017,stuke_chemical_2019} and clustering techniques\cite{de_mapping_2017}.
        
\textbf{QM9 (134k organic molecules):} The QM9 dataset\cite{ramakrishnan_quantum_2014} is a subset of the GDB-17 database \cite{ruddigkeit_enumeration_2012} which was compiled by enumerating all organic molecules that contain up to 17 atoms of C, N, O, S and halogen elements. The QM9 dataset features the first 133,814 molecules from the GDB-17 database. It contains small organic molecules with up to 9 heavy atoms (C, N, O and F), which comprise 621 stoichiometries of small amino acids and nucleobases (pharmaceutically relevant organic building blocks). 
Molecular structures and labels were computed at the PBE+vdW DFT level by Stuke \textit{et al}\cite{stuke_chemical_2019}. 
Despite considerable redundancy, QM9 dataset was used in a variety of ML studies and has become the drosophila of ML in chemistry.

\textbf{OE (64k opto-electronically active molecules):} The OE dataset \cite{stuke_atomic_2020} consists of 64 710 large organic molecules with up to 174 atoms. The structures were extracted from monomolecular organic crystals in the Cambridge Structural Database\cite{allen_cambridge_2002} (CSD) by Schober et al\cite{schober_virtual_2016, schober_ab_2017} for their high charge carrier mobility, and  re-optimized in vacuum at the PBE+vdW level of DFT theory.
OE contains molecules with 16 different elements: H, Li, B, C, N, O, F, Si, P, S, Cl, As, Se, Br, Te, and I.  The molecular structures are more complex than in QM9 and AA, with large conjugated backbones and unusual functional groups. Of the three datasets in this work, OE offers the largest chemical diversity, both in terms of molecule size and number of different elements. OE has become one of the benchmark datasets for molecule generation\cite{westermayr_high-throughput_2023} and property prediction\cite{westermayr_physically_2021, choi_scalable_2022}.

\section{Methodology}
\label{sec:acq-strategy}

To establish the AL approach in this study, we first derived the AL workflow for guided dataset curation and proposed several ASs for picking data. Next, we selected the materials descriptor, ML model and the metrics for evaluating the success of AL. We describe all these steps and their implementation below.  

\begin{figure}[htbp!]
\includegraphics[keepaspectratio,width=\linewidth]{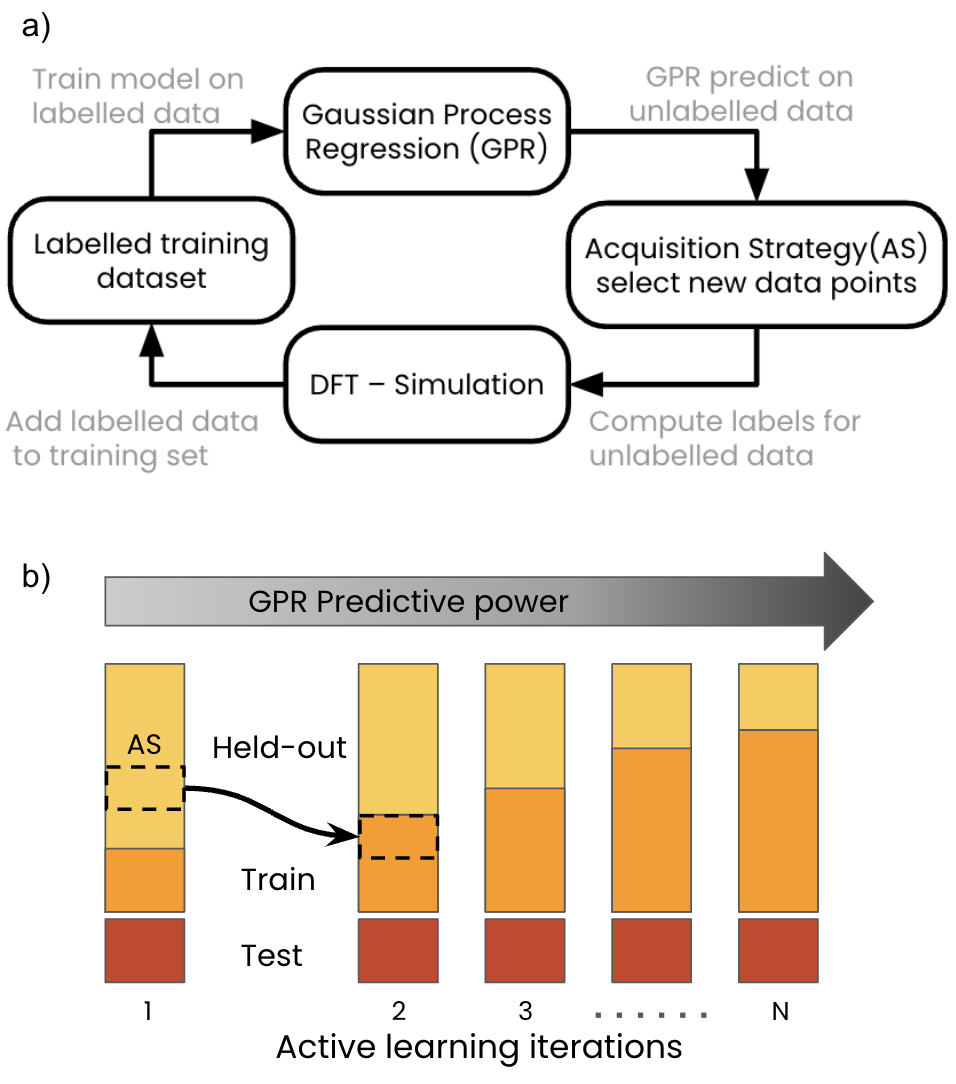}
\caption{\label{fig:main} Illustration of AL steps for a) the active learning iteration and b) the evolution of  held-out, train and test set sizes. Before performing active learning, small, labeled training and test sets are compiled. A Gaussian process regression (GPR) model is fitted to the training set and then used to obtain property predictions of the unlabeled held-out set. The acquisition strategy (AS) combines the predicted property, the corresponding prediction uncertainty and molecular representation, to select molecules from the held-out set. Selected molecules are then labeled using \textit{ab-initio} simulation software (DFT) and added to the training set. The larger training set is used to train a new GPR, and the iteration continues.}
\end{figure}
    
\begin{figure*}[ht!]
\includegraphics[keepaspectratio,width=0.9\linewidth]{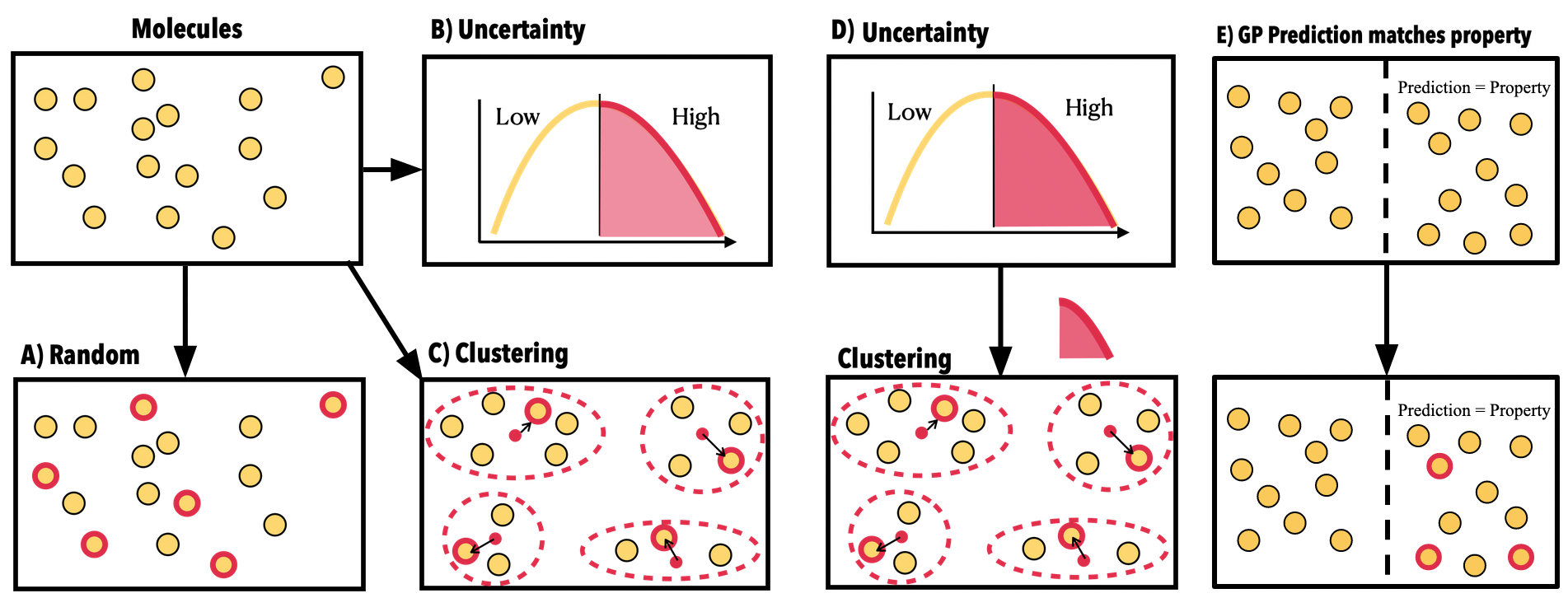}
\caption{\label{fig:strats} Illustration of active learning acquisition strategies: A) random; B) utilizing GPR prediction uncertainty; C) by clustering molecular representations; D) by first selecting molecules with high GPR prediction uncertainty and then clustering the selected molecules, selecting the cluster centers; E) by selecting a set of molecules with GPR predicted property lying within a property value range. Subsequently, a random selection is made from the previous set. The round yellow circles indicate molecules. Round circles, with a thick red border, illustrate selected molecules. Dashed lines separate groups of molecules, red dashed line indicate clusters of molecules. The red dot inside a cluster indicates the cluster center, and the arrow illustrates the molecule closest to the cluster center.}
\end{figure*}

AL molecular dataset curation is the process of assembling the ML training set in an iterative way, by selecting groups of data points from a large pool of molecules called the held-out set. As illustrated in Figure \ref{fig:main}a), each batch would typically be labeled by DFT simulations, before the GPR ML models \cite{rasmussen_gpml_2006} are fit to the training set. GPR outcomes are utilized in ASs that determine how best to select the next batch of data for maximum improvement of ML models, given the objective.  

We encoded this procedure into our AL workflow, depicted in Figure \ref{fig:main}b). Before the start, a batch of molecules was set aside from the held-out set to form the test set, which was kept fixed throughout all AL experiments. The test set served to evaluate the performance of the ML models with the evolving training set. At the start, the first batch of $\mathbf{N}_{\text{init}}$ molecules was selected from the held-out set to nucleate the training set and fit the first GPR model. The model was then applied to predict the property for all the remaining molecular structures in the held-out set. We utilized the prediction results to construct a selection criterion for the AS, typically referred to as the oracle in the AL literature \cite{settles_active_2009}. Consequently, a batch of \nb molecules was selected from the held-out set and appended to the training set. Here, DFT simulations would typically be required to label the molecular structures, but we expedited the tests with pre-computed labels. A fully labeled augmented training set was the outcome of a single AL iteration. The next AL iteration began by retraining the GPR with the updated training set.

The quality of the curated dataset depends on how intelligently an AS can pick molecules from the held-out set. The AL literature is rife with various AS designs, for example, strategies for compiling thin-film materials \cite{wang_benchmarking_2022}, potentials for metal-organics\cite{vandenhaute_machine_2023}, or the design of layered materials\cite{bassman_oftelie_active_2018}. While these strategies addressed specific tasks (task 2 here), it is also important to consider AS designs for general ML accuracy with minimal datasets (task 1 here).

The selection strategy in acquisition functions is typically based on a trade-off between diversity and redundancy. Increasing diversity ensures good representation of different kinds of molecules and reflects data space exploration. Higher redundancy allows models to learn minor variations in property values of similar molecules, through data exploitation. In practice, the trade-off is implemented through considerations of GP prediction uncertainty and clustering algorithms. When GP models are applied to the molecular structures in the held-out set, prediction uncertainty is highest for the structures that differ most from the molecules in the training set. Uncertainty-based picking thus increases the diversity of the training set. However, molecules with high prediction uncertainty could all be similar to each other. Similarity among the selected structures can be minimized by clustering them and choosing data from each cluster. The number of clusters correlates with the diversity of the selection, and the number of molecules selected from each cluster determines the redundancy. 

While all these considerations are relevant, it is unclear which combination of AS choices and related parameters would lead to most accurate ML models, trained on most compact, maximally informative datasets. To address this question, we constructed five AS and compared them in the AL workflows described above. The following AS selection rules are illustrated in Figure \ref{fig:strats}).

\textbf{(A) Random:} In the simplest strategy, we randomly selected a fixed number of molecules from the held-out set, labeled them and added the data to the training set. Such sampling ensures even representation of data across the held-out set, but also encodes all biases and redundancies. This traditional sampling strategy often yields accurate machine learning models\cite{reker_2020}. It served as the baseline against which we compared AL approaches.

\textbf{(B) Uncertainty:} This strategy utilized GP prediction uncertainty to add molecules to the training set. At each iteration, after training the GP, we computed the predictions and corresponding uncertainty values on the held-out set molecules. Data indices were sorted based on their prediction uncertainty, and a batch of $\mathbf{N}_b$ molecules with the highest uncertainty were selected. This AS encourages exploration and leads to rapid ML uncertainty reduction, without any considerations of diversity.

\textbf{(C) Clustering:} Molecules in the held-out set were grouped into $\mathbf{N}_b$ clusters. From each group, one structure closest to the cluster center was selected and added to the training set. The objective here was to maximize the structural diversity of molecules in the training set, irrespective of uncertainty. 

\textbf{(D) Uncertainty and Clustering:} 
The trained GP model was applied to compute predictions on the held-out set, after which the data was sorted by GP prediction uncertainty. The top 50\% of the molecules with highest uncertainty were selected and divided into $\mathbf{N}_b$ clusters. Molecules closest to cluster centers were added to the training set. This AS combines the two previous ones to overcome their respective shortcomings. 

\textbf{(E) Property search:} This AS was specifically designed for task 2. The trained GP model was applied to compute property predictions on the held-out set, after which the molecular structures were filtered based on the target property criterion (here, HOMO > $\varepsilon$). From all the molecules with this predicted property, $\mathbf{N}_b$ structures were chosen at random and added to the training set. In the early stages of AL, poorer GP model accuracy may lead to less accurate selections. As iterations proceed, more molecules matching the search criterion should be found, emulating data exploitation.

The key AS parameter is \nb, the number of molecules added to the training set with each iteration. \nb can be fixed (e.g. $\mathbf{N}_{\text{const}}$=1000 or 1k molecules) or adaptive, evolving with each iteration. Adaptive \nb could prove important because GP models may have limited accuracy in the early iterations of active learning, favoring small \nb. Nevertheless, small batches would be inefficient in the later stages, when model accuracy improves and larger batches are better suited.
We implemented a power law (POW) batch scheme to evolve \nb with iteration $t$ as ${N}_{b}(t)$=$2^{t} * \mathbf{N}_{\text{const}}$, where $\mathbf{N}_{\text{const}}$=1000. 
This meant that the batch size doubled in size with each AL iteration $t$, making the training set size 
$\mathbf{N}_{TR}(t)$:
\begin{equation}
\mathbf{N}_{TR}(t) = 2^{t} * \mathbf{N}_{\text{const}} + \mathbf{N}_{\text{init}} \qquad \forall \; t \in \mathbb{N}.
\label{eq:pow}
\end{equation}

We also tested a constant batch scheme, where $\mathbf{N}_{\text{const}}$ was either 1k, 2k, 4k or 8k: 
\begin{equation}
\mathbf{N}_{TR}(t) = t * \mathbf{N}_{\text{const}} + \mathbf{N}_{\text{init}}  \qquad \forall \; t \in \mathbb{N}.
\label{eq:const}
\end{equation}

The initial batch size of $\mathbf{N}_{\text{init}}$ was set to 1000 molecules, and remained fixed for all increment schemes. The indices of molecules in the initial batch were kept exactly the same while computing the learning curves to ensure comparable results.

Molecular structures were encoded with the many-body tensor representation (MBTR) \cite{huo_unified_2018}, which has demonstrated superior accuracy in ML studies on molecular datasets\cite{rahaman_2020, bahlke_2020, lumiaro_2021}.
The MBTR records atomic species, pairwise distances and angles between atoms as components of a vector denoted by $k1$, $k2$ and $k3$ respectively. Previous work\cite{stuke_chemical_2019} has shown that omitting terms $k1$ and $k3$ results in a small loss in accuracy with molecular datasets, but a large reduction in descriptor size. Consequently, we restricted our MBTR vectors to $k2$ terms only. For details on MBTR hyperparameter optimization, we refer the reader to our previous work\cite{stuke_chemical_2019}. 

Since the MBTR vectors vary smoothly, they can be modeled well with GPs in kernel-based supervised learning regression.
These non-parametric models access the entire training data in the form of a kernel matrix for the model to learn. While GPR does not scale very well to large training datasets, it does provide an intuitive means of obtaining prediction uncertainty, required for the AL strategies. 

We trained the GP to predict  molecular HOMO energy levels (target $\mathbf{y}$) based on the molecular representations (input $\mathbf{X}$). The trained model was then applied to previously unseen molecular structures  $\mathbf{X_{*}}$, to compute the posterior predictive mean $\mu_{\text{pred}}$ (HOMO level prediction) and posterior predictive variance or uncertainty $\sigma^{2}_{\text{pred}}$. Given a data noise of $\sigma_{n}^2$, 

\begin{equation}
\mu_{\text{pred}} = K(\mathbf{X_{*}}, \mathbf{X}) \left[ K(\mathbf{X}, \mathbf{X}) + \sigma_{n}^2\mathbf{I} \right]^{-1}\mathbf{y}\\
    \label{eq:pred_mean}
\end{equation}

\begin{equation}
\sigma^{2}_{\text{pred}} = K(\mathbf{X_{*}}, \mathbf{X_{*}}) - K(\mathbf{X_{*}}, \mathbf{X}) K_{x}^{-1} K(\mathbf{X}, \mathbf{X_{*}})
    \label{eq:pred_var}
\end{equation}

\noindent In Equations \ref{eq:pred_mean} and \ref{eq:pred_var} the symmetric kernel matrix  $K$  has individual matrix elements computed through the kernel function. Because the MBTR vector is smoothly varying, we selected the radial basis function (RBF) kernel ($k_{RBF}$):

 \begin{equation}
    K_{x} = K(\mathbf{X}, \mathbf{X}) + \sigma_{n}^2\mathbf{I}
    \label{eq:kernel}
\end{equation}

\begin{equation}
\begin{split}
    k_{RBF}(x,x') &= \sigma_{f} \ \text{exp}\left( {\frac{-{\lVert x - x' \rVert}^2_2 }{2l^2}} \right) \\
\end{split} \label{eq:sqexp_def}
\end{equation}

In Equation \ref{eq:sqexp_def}, the exponent can be expanded as follows:
\begin{equation}
    {\lVert x - x' \rVert}_2 = \sqrt{ \sum_i(x_i - x'_i)^2 }
\end{equation}

This is the Euclidean distance between the two MBTR vectors $\mathbf{x}$ and $\mathbf{x'}$. In Equation \ref{eq:sqexp_def}, $\sigma_{f}$ refers to a global scaling factor or signal variance in the GP literature\cite{rasmussen_gpml_2006}, and $l$ represents the length scale. 

\begin{figure*}[t]
\includegraphics[width=\linewidth]{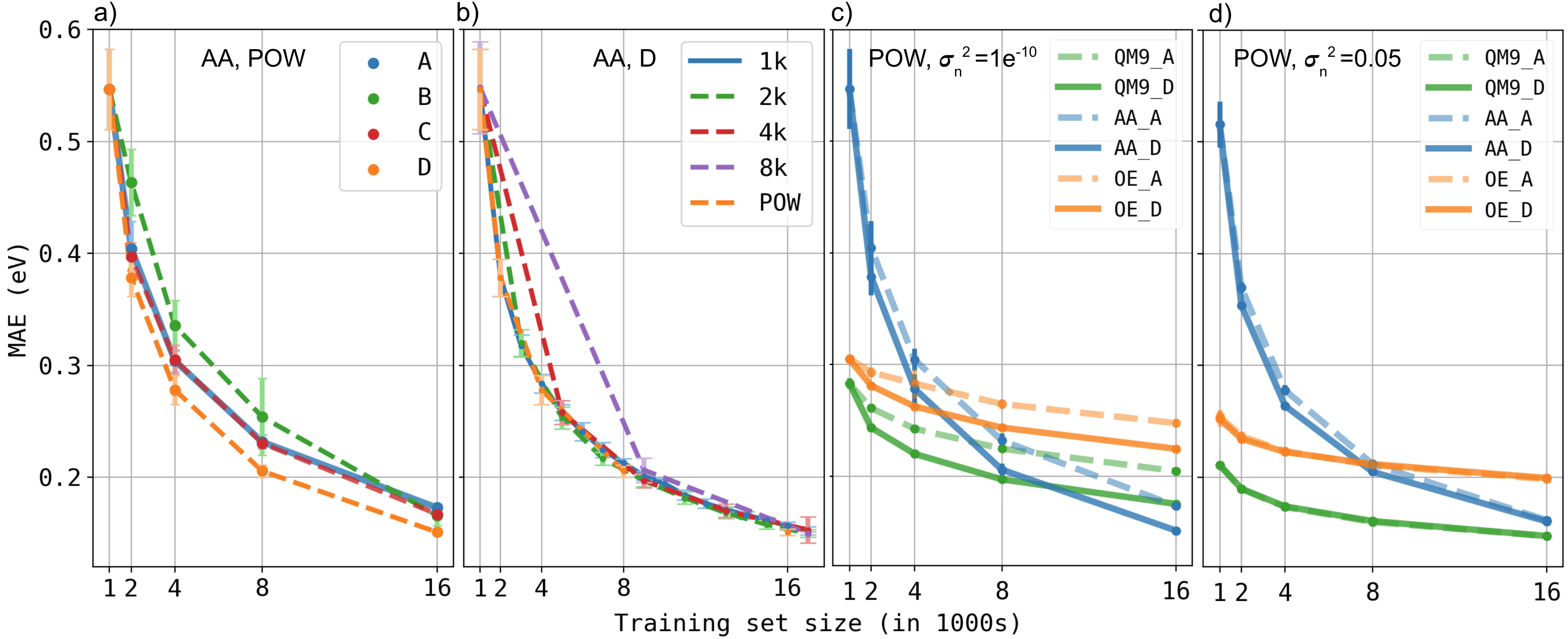}
\caption{
AL learning curves for Task 1, with test set MAEs computed from GP model predictions as a function of increasing training set size. a) Performance of different AS for the AA dataset with the POW batch scheme. b) Performance of different batch strategies for the AA dataset and AS D. c) Strategy A and D performance on all datasets with  $\sigma_n^{2}$=$10^{-10}$. d) Strategy A and D performance on all datasets with  $\sigma_n^{2}$=0.05. }
\label{fig:strat-bs-noise}
\end{figure*}

We implemented the GPR in the Scikit-learn\cite{scikit-learn}  (SKLearn) package and optimized the model hyperparameters by maximizing the \textit{log marginal likelihood}\cite{rasmussen_gpml_2006}, which is a function of the training labels $\mathbf{y}$ and GP mean and variance computed on the training data ($\mu$ and $\sigma^{2}$ computed with $\mathbf{X}_{*}=\mathbf{X}$ in Eqs. \ref{eq:pred_mean} and \ref{eq:pred_var}). The length scale and output variance were initialized to 700 and 20 to match the hyperparameters from previously published\cite{stuke_chemical_2019} ML models trained on the same datasets. The hyperparameters were optimized over a range which varied four orders of magnitude, with the lower and upper hyperparameter search bound set 100 times smaller or larger than the initial value. 
The GP prior mean was set to zero (\texttt{normalize\_y = False} in SKLearn) and the number of optimizer restarts were set to 2. Given that the data labels were obtained with accurate DFT simulations, we set the model noise to a very low value of $\sigma_n^{2}$=$10^{-10}$ (unless otherwise stated).

To evaluate GPR model performance, we used both regression and classification metrics. 
To monitor regression, we computed the mean absolute error (MAE) of the model on the test set as a function of training set size. This allowed us to build ML learning curves and compare different AS approaches. 
In task 2, we gained further insight into the classification accuracy, via the following classification metrics. \cite{fawcett_introduction_2006} 
TPR is the ratio of the number of molecules correctly classified to be in-range (\textit{true positive} or TP) and the total number of molecules in this class (\textit{positives} or P). Similarly, FPR is the ratio of the number of molecules incorrectly classified to be in-range (false negative or FN) and the number of molecules which are not in the correct class (\textit{negatives} or N). The TPR and FPR values range between 0 and 1. As the classification accuracy of the ML model improves, TPR tends towards 1 and FPR towards 0.
\begin{equation}
    \begin{split}
    \text{TPR} = \frac{TP}{P} \\
    \text{FPR} = \frac{FN}{N}. \\
    \end{split}
    \label{eq:TPR-FPR}
\end{equation}


\section{Results}
\label{sec:results}

We began by identifying the best performing AS to compile maximally compact and informative datasets, as defined in task 1. The relative performance of different acquisition strategies (AS) was evaluated by comparing their learning curves. We explored the efficacy of different AS as a function of key method parameters: the acquisition batch size \nb which affects the growth rate of the training set and the estimated data noise $\sigma_n^{2}$ in the GP model, which determines the smoothness of the model and consequently its performance. We also considered how the proposed ASs perform on different molecular datasets.

The first series of AS tests was carried out with the AA dataset, because it was structurally the most redundant one. In our bid to actively learn the most compact training set, there was considerable similarity to be eliminated from the pool of AA molecules, and we expected to observe the largest difference in performance between the proposed AS schemes. Figure \ref{fig:strat-bs-noise}a) presents the AS learning curves (averaged over 5 runs) computed with the POW \nb incrementing scheme for task 1, as the training set was curated from 1k to 16k molecules. Uncertainty-based strategy B was the worst AS, consistently achieving prediction errors higher than the random picking baseline A. At maximum training set size, we observed $\mathrm{MAE_B}$=0.164eV$\pm0.001$ compared to $\mathrm{MAE_A}$=0.170eV$\pm0.009$. AS C, which clusters molecules based on their structural similarity, performed on par with the baseline, achieving an $\mathrm{MAE_C}$=0.166eV$\pm0.001$. Strategy D which combines uncertainty and clustering performed the best with $\mathrm{MAE_D}$=0.148eV$\pm0.002$. It was selected for the next set of tests.

Batch sampling selects multiple molecules at a time, and while it allows AL to scale to larger datasets\cite{citovsky_neurips_2021}, it is more susceptible to sampling redundant molecules. This is why we tested the sensitivity of strategy D to \nb choice. The POW incrementing scheme was compared to the constant \nb approach, with $\mathbf{N}_{\text{const}}$ of 1k, 2k, 4k or 8k. Results are presented in Figure \ref{fig:strat-bs-noise}b).  Surprisingly, all sampling schemes exhibited the same ML performance. However, smaller constant \nb resulted in more AL iterations and were consequently slower to execute. We selected the POW scheme to proceed with because it covers a wide range of training set sizes with the fewest AL iterations.

Next, we considered if strategy D performs equally well for different datasets. Figure \ref{fig:strat-bs-noise}c) illustrates the learning curves for strategy D against random picking for datasets AA, QM9 and OE. The learning rates for the three datasets were very different, as indicated by our previous study \cite{stuke_chemical_2019}. In all cases, strategy D appeared to lower MAEs compared to the baseline, suggesting that similar learning could be achieved with a smaller training set size. However, the MAEs for the baseline strategy A were consistently higher compared to the same data from previous work \cite{stuke_chemical_2019}, alerting us to a problem. 

A careful review of our GPR revealed that our data noise settings $\sigma_n^2 = 10^{-10}$  was quite low and lead to over-fitting. We conducted a grid search for optimal GP noise and found it to be as high as $\sigma_n^2 = 0.05$. To establish how data noise settings affect the performance of the AL, we recomputed the learning curves of strategies D and A for all datasets with the optimal $\sigma_n^2$ levels. The results in Figure \ref{fig:strat-bs-noise}d) now indicate that at higher noise levels, the performance of strategy D is indistinguishable from random picking A. Apart from minor savings for the redundant AA dataset, we discovered no benefit of active learning for task 1.

Next we focus on Task 2, where the objective is to identify molecules with HOMO energy greater than $\varepsilon$, merely based on structure. $\varepsilon$ was chosen such that, approximately 30\% of the molecules in the datasets belong to this category. This resulted in $\varepsilon_{QM9} = -5.55$eV , $\varepsilon_{AA} = -8.5$eV and $\varepsilon_{OE} = -5.2$eV. Here, we inspect the performance of Strategy E against Strategy A for all data sets, with the POW batch scheme and optimal $\sigma_n^2 = 0.05$. Since this is a classification task, model performance can be evaluated by how many molecules of the correct class (HOMO in-range) are extracted by the AS from the total structural pool, verified against the computed HOMOs. 

\begin{figure}[htbp!]
\includegraphics[width=\linewidth]{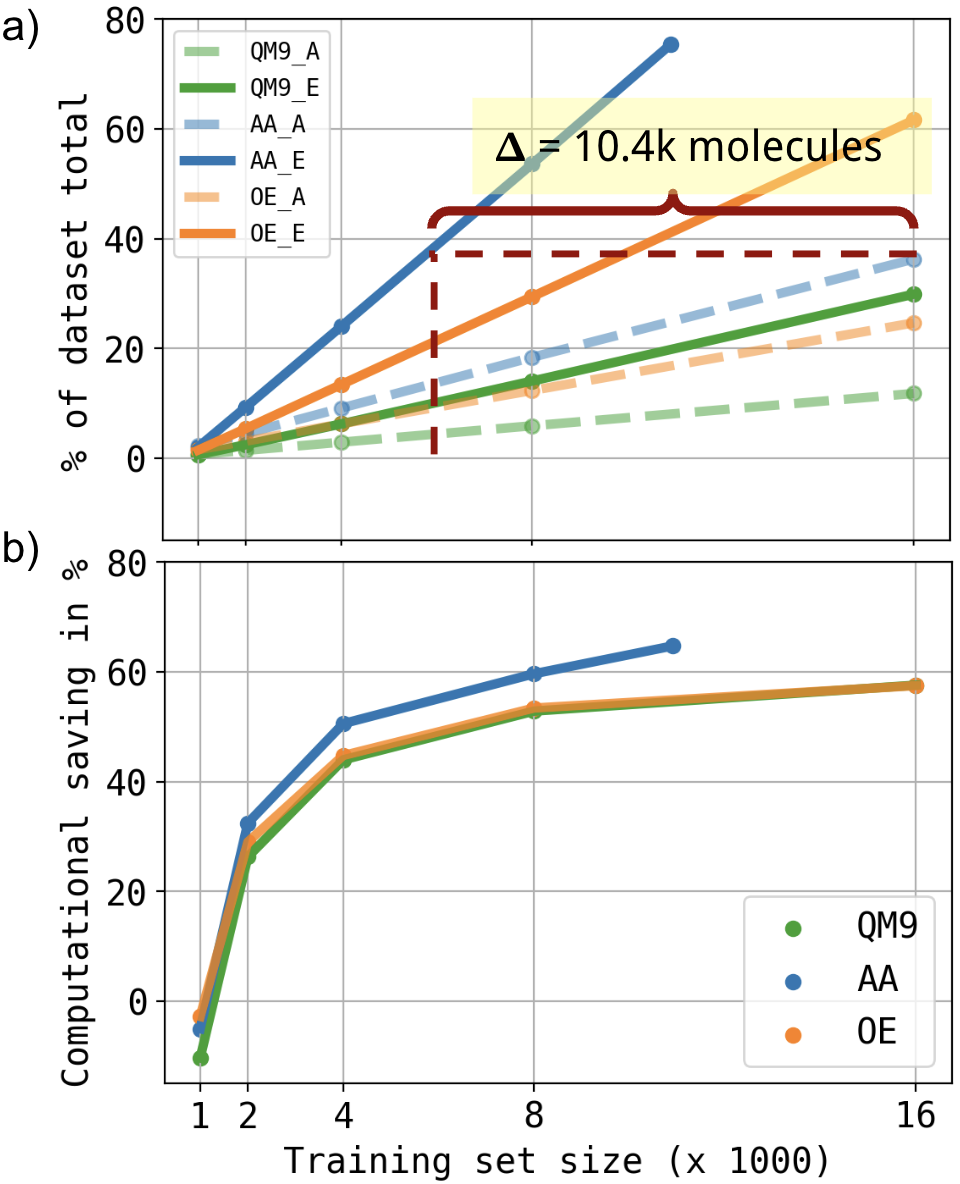}
\caption{AL model performance for Task 2. a) Number of correct structures (HOMO > $\varepsilon$) identified by AS A and E, presented as a percentage
of the total in-range molecules in the dataset. b) 
As seen in panel a), AS E requires fewer training examples to achieve the same predictive accuracy as AS A. The plot presents the number of additional in-range molecules identified by AS E relative to AS A, expressed as a percentage of total in-range molecules in each dataset.
}
\label{fig:datasaving-pc}
\end{figure}

The learning curves in Figure \ref{fig:datasaving-pc}.a illustrate the success of molecular classification as a function of GP model training set size. Since QM9, AA and OE have different dataset sizes, the number of correct structures is presented as a percentage of the total in-range molecules in the dataset. Model performance increases linearly with the training set size, with a larger slope indicating a faster rate of extracting in-range molecules. 

It is evident that AS E achieves systematically higher identification rates and extracts more correct structures compared to the random baseline. Different learning rates indicate that the molecules in the AA datasets were easiest to classify (over 70\% correctly identified), followed by OE and QM9. 
Better AS E performance against the random baseline means that the same number of correct molecules could be identified in fewer iterations. For AA, 39\% of the in-range molecules were identified with a training set size of 5600 with strategy E (dotted line), and the same percentage was achieved by the baseline with 16000 training molecules. Less training set data translates to 10,400 fewer HOMO computations for the same model quality. We used the two learning curves to calculate the fraction of relative computational savings produced by strategic sampling. The data is illustrated in Figure \ref{fig:datasaving-pc}b). Best savings could be obtained for the AA dataset (64\%), while the savings for OE and QM9 were surprisingly similar (57\%).

\begin{figure}[t!]
     \centering
     \includegraphics[width=\linewidth]{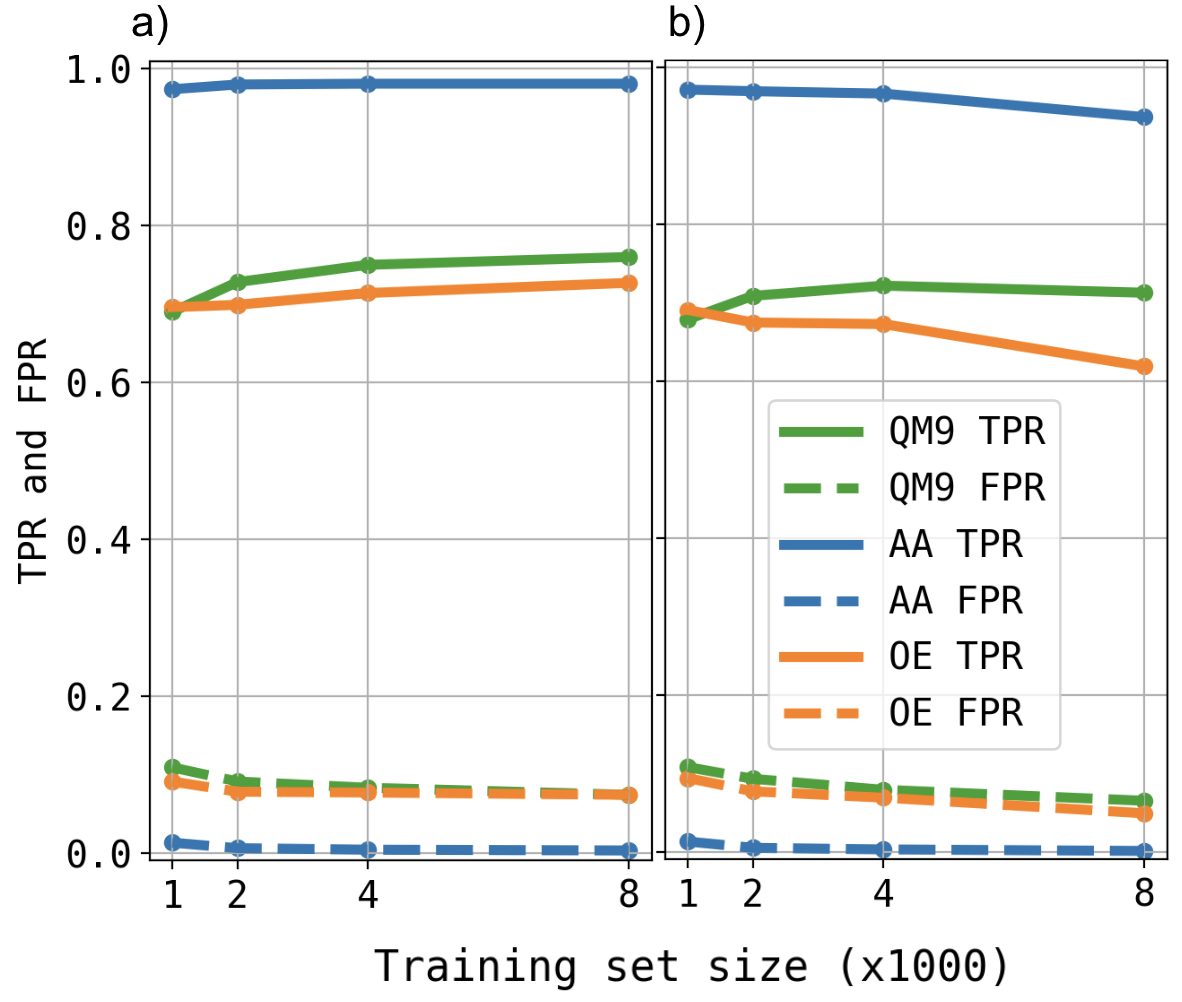}
     \caption{Task 2 classification metrics TPR and FPR as a function of training set size, evaluated with strategy E for all 3 datasets on the a) test set and b) held-out set.}
     \label{fig:classification-metrics}
 \end{figure}

Next, we inspect the classification accuracy of AS E. Figure \ref{fig:classification-metrics} illustrates the rate of TP and FP classifications as a function of training set size. The metrics computed on a randomly chosen test set in Figure \ref{fig:classification-metrics}a), assesses the objective improvement in classification accuracy. 
It is interesting to observe that the model classifies well already with little training data. The improvement in TP with more data is very slight, almost none for the AA dataset. The FP wrong classifications register a very small decrease on the test set. 

We also explore the metrics on the held-out set in Figure \ref{fig:classification-metrics}b) to establish if the classification accuracy changes as the held-out set is depleted of relevant molecules with subsequent iterations. There is a slight overall decrease in accuracy, which suggests that as more relevant molecules are moved from the held-out set to the training set, the model cannot classify the remaining structures as well. This trend is most notable in TP data for the AA and OE datasets, where structural diversity is high. The drop in accuracy leads to fewer relevant molecules added to the training set in later stages of AL. 

\begin{figure*}[bth!]
\includegraphics[keepaspectratio,height=1.1\linewidth, width=\linewidth]{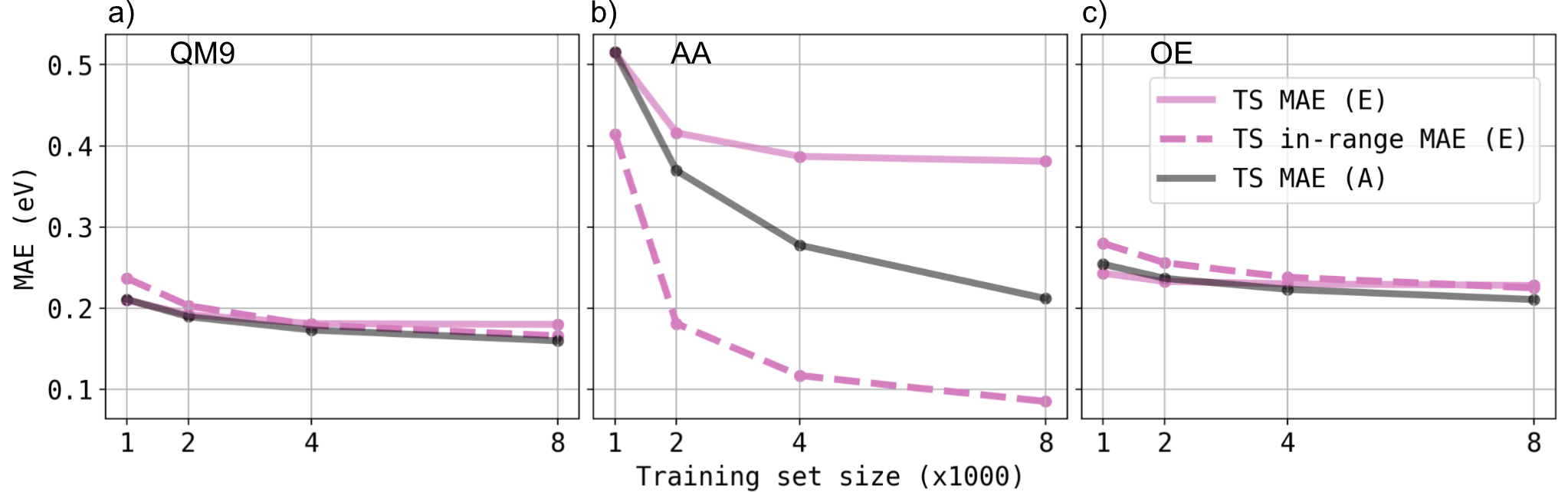}
\caption{Test 2 regression MAE metrics evaluated on a) QM9, b) AA and c) OE datasets. MAE evaluated with AS E on the test set (pink solid line) and a test set (TS) of in-range molecules (pink dashed line) are compared with those evaluated with AS A on the test set (black solid line).
}
\label{fig:testset-MAE}
\end{figure*}

In this task, classification accuracy of the model acutely depends on the quality of supervised regression: the prediction of the HOMO energy based on molecular structure. Figure \ref{fig:testset-MAE} presents the evolution of prediction MAE with training set size for all datasets. Overall, we observe the prediction errors decreasing, as expected. The results are most notable for the AA dataset in Figure \ref{fig:testset-MAE}b). Here, the prediction MAEs on the in-range molecules are reduced dramatically compared to the errors on the random subset. This is not the case for the QM9 and OE datasets, where MAE of predictions on the in-range and randomly selected molecules are nearly identical.

\section{Discussion}

Among the two proposed active learning objectives, Task 1 is the most commonly studied in the literature\cite{zhou_jointly_2022, hwang_uncertainty-based_2022,kulichenko_uncertainty-driven_2023, zhang_bayesian_2019}.
We carried out a comparative study of AL strategies to explore which selection criteria curate superior training sets for best predictive models (see Figure \ref{fig:strat-bs-noise}a)).  We observed that combining prediction uncertainty with clustering molecular structures (Strategy D) yields the best results (lowest test set MAE), which corroborates similar findings in the literature\cite{zhou_jointly_2022, hwang_uncertainty-based_2022}.

In this study, uncertainty alone was a sub-optimal criterion and consistently yielded higher MAEs compared to all other strategies. This, however, contradicts other findings in the literature \cite{kulichenko_uncertainty-driven_2023, zhang_bayesian_2019}. 
Such a discrepancy can be explained by two key differences: the choice of machine learning model and the optimization task.
In this GP-based study, uncertainty is modeled  differently\cite{li_deep_2021} than with the neural networks employed in earlier work\cite{zhang_bayesian_2019}.
Similarly, the focus here is on predicting HOMO energy value instead of inter-atomic potentials\cite{kulichenko_uncertainty-driven_2023}. Both these factors could lead to different learning outcomes, a factor to which we will return later in the discussion section.

Despite previous work with different batch sizes in AL\cite{zaverkin_exploring_2022}, there is no insight into how batch size affects the quality of the GP models. 
Here, we compare constant and adaptive batch schemes and observed that batch size has negligible effect on the performance of the best acquisition strategy.

In contrast, dataset noise ($\sigma_n$) has a significant impact on the performance of AL. 
If the GP overfits on the training data for low $\sigma_n$ values, the test set MAE increases, as observed for the random strategy (see AS A, in panels c and d of Figure \ref{fig:strat-bs-noise}). AL can then provide a benefit by balancing the dataset through clustered uncertainty minimization (compare AS A and D in Figure \ref{fig:strat-bs-noise}c). Strategy D thus ensures a higher diversity than random sampling, which reduces overfitting. The GPR of strategy D subsequently generalizes better to the test set and the MAE drops faster. For larger noise values, the accuracy of the GP increases as overfitting on the training set reduces, as observed in Figure 4d. The AL benefit disappears, since the best strategy to resemble the randomly drawn test set also in the training set, is to randomly assemble it.

There are conflicting reports in the literature on the benefits of AL in dataset curation. However, a closer look reveals that benefits depend on the prediction task, and more specifically on the bounds of the search space sampled by AL. Training interatomic potentials is a prototypical example of an unbounded search task, where atomic configurations are sampled from a near-infinite pool of possible structures. Since atoms in these tasks can explore real-space continuously, there is no limit to the number of possible structures to pick.
The test set, however, is constrained to a certain part of real space governed by the laws of quantum mechanics, namely the vicinity of equilibrium structures (interatomic potentials) or the surroundings of a molecular dynamics trajectory. This generates an intrinsic difference between the distributions of structures in the training and test set.  AL can then efficiently reduce the number of training structures as it aligns the training with the test set\cite{zaverkin_exploration_2021}. 

\begin{figure}[hb]
\includegraphics[width=\linewidth]{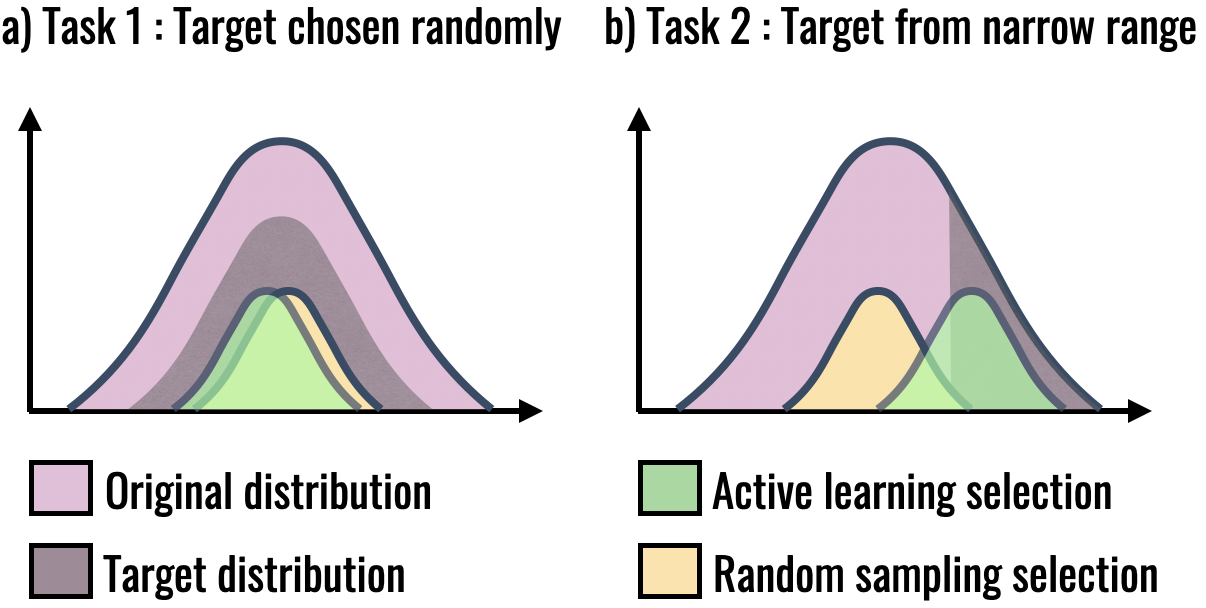}
\caption{Illustration of the two active learning tasks. a) In task 1, the target is selected randomly from entire original distribution, consequently random sampling achieves a good representation of the target and active learning provides no benefit over random sampling. b) In task 2, the target is selected from a narrow region of the original distribution, here active learning can adapt and represent the target well, outperforming random sampling.} 
\label{fig:twotasks}
\end{figure}

This work deals with dataset curation from a large but finite pool of molecular structures. This is standard for property prediction tasks, in the absence of generative models to open up the search space bounds. Here, mere random picking already generates a matching distribution of the training and test sets, as illustrated in Figure \ref{fig:twotasks}a). AL sampling can hardly offer any benefit in re-sampling the training set, since there is no difference to bridge. Our finding in task 1 is therefore consistent with previous work\cite{zaverkin_exploring_2022}. In task 2, however, targeting molecules with particular HOMO values translates to a target distribution shifted away from the general structure pool (Figure \ref{fig:twotasks}b)). The HOMO > $\varepsilon$ target constitutes the upper third of the label distribution, which would be sampled poorly by the random strategy A. For such a use case, AL quickly focuses on the appropriate structures and provides a distinct benefit.
It follows that the benefits of AL can be observed only in the case of a shift between the structural distributions of the test set and the overall search space. When the structure pool is finite, test set engineering can be used to fabricate a shift.

Targeted property search presents an example of intrinsic test set engineering, where AL can yield computational benefits. Strategy E was able to identify more desirable molecules in fewer AL iterations, in comparison to random selection.
This occurs because strategy E biased the training set with more molecules in the correct HOMO category, which improves predictive accuracy of the model in the target HOMO range. Such a strategy of biasing the training set can be detrimental to the learning process\cite{richards_2011, zhang_et-_2023, farquhar_2021}. However, when used judiciously for a specific task, biasing the training data has been shown to improve model performance\cite{filstroff_targeted_2021, westermayr_high-throughput_2023, sundin_active_2019,besel_2024}.

We observed that AL benefits can vary considerably with dataset type. The magnitude of computational savings identified in Figure \ref{fig:datasaving-pc} arises directly from the classification accuracy observed in Figure \ref{fig:classification-metrics}. For the AA dataset, near-perfect classification was observed even for small training sets. Accuracy for the QM9 and OE datasets improved slowly but converged to a constant value below 0.8 in TPR. To interpret this, we review the distribution of HOMO labels in Figure \ref{fig:homo-dist} of the Appendix with regard to the position of the $\varepsilon$ property boundary for correct classification. QM9 and OE boast an unimodal HOMO distribution. Since the decision boundary is placed near the distribution mode, many molecular structures are vulnerable to misclassification already for small HOMO prediction errors. This ultimately limits the classification accuracy, even at large training sets.
In the bimodal distribution of HOMOs for the AA dataset, the decision boundary includes much of the relevant peak, which is why the initial accuracy is high. The subsequent addition of structures from the second mode of low HOMO values cannot improve on this classification training set. AA classification accuracy is therefore consistently high, and that translates to good selectivity and large-scale computational savings.

The position of the decision boundary within the HOMO distribution also explains the quality of supervised regression in Figure ~\ref{fig:testset-MAE}. For QM9 and OE datasets, the boundary near the HOMO label mode means that the molecules on either side of the classification boundary are similar. The randomly drawn dataset and the AL dataset capture similar information. For the bi-modal distribution of AA HOMO values, the AL model is primarily built on molecules from one mode of the distribution, whereas the random model contains both. Since the lower HOMO peak contributes little useful structural information to the model, AL performance is notably better.

As the GPR models grow more accurate, it is interesting that the classification accuracy on the held-out set molecules in Figure ~\ref{fig:classification-metrics}.b) is reduced for all datasets. This indicates that successive AL iterations deplete the held-out set of molecules that are relevant and easy to classify. 
The GPR model accuracy is improved, but because the remaining structures are harder to classify, the net effect is the decrease of TPR for all three datasets.

\section{Conclusion}

In this study, we proposed novel AL methodology for applications in molecular and materials science. The objectives were to curate compact and maximally informative datasets, or identify molecules with targeted properties with the fewest calculations performed. The performance of proposed algorithms was analyzed to identify the best settings to employ AL. Our results revealed that, for finite size datasets deployed in this work, AL provides no benefits for minimizing global error metrics such as the MAE. Instead, we found that computational savings achieved with AL are dependent on the distribution of target molecules in the task, with respect to the total dataset distribution. These observations help to reconcile seemingly contradictory reports in the AL literature. 

For applications minimizing global errors such as MAE, the target distribution is drawn randomly and resembles the parent distribution. Here, the best AL strategy is to draw randomly, instead of criteria based on uncertainty and diversity. Our findings indicate that AL provides significant computational savings in applications where the target molecules are drawn from a narrow region of dataset. For targeted property search, the proposed AL strategy provided significant computational savings, of up to 64\% as compared to random sampling. The savings can vary considerably with the compound space sampled. Searching for molecular structures with targeted properties therefore presents a useful application in materials research, where AL delivers the most benefit.

\begin{acknowledgments}
This study received financial support from the Academy of Finland through its flagship program, the Finnish Center for Artificial Intelligence and the Centers of Excellence Program (CoE VILMA, Grant No. 346377).
Computing resources from the Aalto Science-IT project and the CSC – IT Center for Science, Finland, are gratefully acknowledged. Additionally, K.G. thanks Finnish Cultural Foundation (Grant No. 00210309) for funding the research.

\end{acknowledgments}

\section*{Author Declarations}

\subsection*{Author Contributions:} MT and PR conceived the initial idea. KG implemented the machine learning model and ran the experiments. All the authors contributed to writing the manuscript.
\subsection*{Ethics Approval} Ethics approval not required.
\subsection*{Conflict of Interest:} Authors claim no conflict of interest.

\section*{Code and Data Availability Statement}

The data that support the findings of
this study are openly available in Zenodo, record numbers \href{https://zenodo.org/record/3967308}{3967308}\cite{ghosh_2020_3967308}, \href{https://zenodo.org/record/4035923}{4035923}\cite{ghosh_2020_4035923} and \href{https://zenodo.org/record/4035918}{4035918}\cite{ghosh_2020_4035918}. The code to reproduce the results in this publication can be found on GitHub\footnote{\url{https://github.com/kunalghosh/Multi_Fidelity_Prediction_GP/tree/testing_runs}}.

\appendix

\section{Distribution of HOMO energy values.}
\label{sec:homo-dist}
\begin{figure*}
    \centering
    \includegraphics[width=\linewidth]{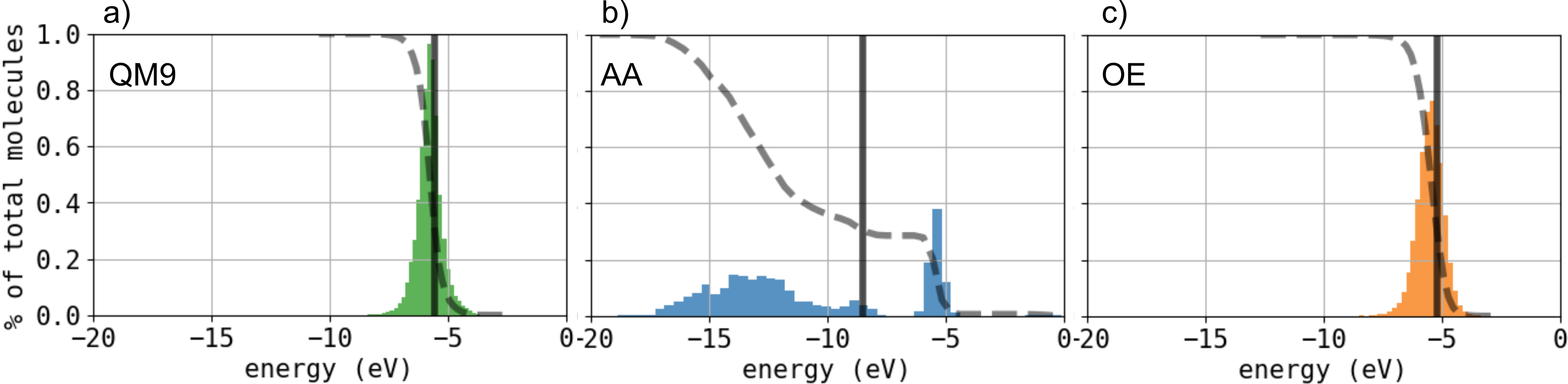}
    \caption{Distribution of HOMO energy values for a) QM9 b) AA and c) OE. The vertical line indicates the classification boundary ($\varepsilon$), and the dashed line indicates the cumulative sum of the distribution of molecules. Classification boundary roughly includes 30\% of the molecules from the dataset. This includes molecules with HOMO greater than -5.55 eV, -8.5 eV and -5.2 eV for QM9, AA and OE respectively.}
    \label{fig:homo-dist}
\end{figure*}

\section*{References}
\bibliography{main}

\begin{thebibliography}{62}%
\makeatletter
\providecommand \@ifxundefined [1]{%
 \@ifx{#1\undefined}
}%
\providecommand \@ifnum [1]{%
 \ifnum #1\expandafter \@firstoftwo
 \else \expandafter \@secondoftwo
 \fi
}%
\providecommand \@ifx [1]{%
 \ifx #1\expandafter \@firstoftwo
 \else \expandafter \@secondoftwo
 \fi
}%
\providecommand \natexlab [1]{#1}%
\providecommand \enquote  [1]{``#1''}%
\providecommand \bibnamefont  [1]{#1}%
\providecommand \bibfnamefont [1]{#1}%
\providecommand \citenamefont [1]{#1}%
\providecommand \href@noop [0]{\@secondoftwo}%
\providecommand \href [0]{\begingroup \@sanitize@url \@href}%
\providecommand \@href[1]{\@@startlink{#1}\@@href}%
\providecommand \@@href[1]{\endgroup#1\@@endlink}%
\providecommand \@sanitize@url [0]{\catcode `\\12\catcode `\$12\catcode
  `\&12\catcode `\#12\catcode `\^12\catcode `\_12\catcode `\%12\relax}%
\providecommand \@@startlink[1]{}%
\providecommand \@@endlink[0]{}%
\providecommand \url  [0]{\begingroup\@sanitize@url \@url }%
\providecommand \@url [1]{\endgroup\@href {#1}{\urlprefix }}%
\providecommand \urlprefix  [0]{URL }%
\providecommand \Eprint [0]{\href }%
\providecommand \doibase [0]{https://doi.org/}%
\providecommand \selectlanguage [0]{\@gobble}%
\providecommand \bibinfo  [0]{\@secondoftwo}%
\providecommand \bibfield  [0]{\@secondoftwo}%
\providecommand \translation [1]{[#1]}%
\providecommand \BibitemOpen [0]{}%
\providecommand \bibitemStop [0]{}%
\providecommand \bibitemNoStop [0]{.\EOS\space}%
\providecommand \EOS [0]{\spacefactor3000\relax}%
\providecommand \BibitemShut  [1]{\csname bibitem#1\endcsname}%
\let\auto@bib@innerbib\@empty
\bibitem [{\citenamefont {Kulik}\ \emph {et~al.}(2022)\citenamefont {Kulik},
  \citenamefont {Hammerschmidt}, \citenamefont {Schmidt}, \citenamefont
  {Botti}, \citenamefont {Marques}, \citenamefont {Boley}, \citenamefont
  {Scheffler}, \citenamefont {Todorović}, \citenamefont {Rinke}, \citenamefont
  {Oses}, \citenamefont {Smolyanyuk}, \citenamefont {Curtarolo}, \citenamefont
  {Tkatchenko}, \citenamefont {Bartok}, \citenamefont {Manzhos}, \citenamefont
  {Ihara}, \citenamefont {Carrington}, \citenamefont {Behler}, \citenamefont
  {Isayev}, \citenamefont {Veit}, \citenamefont {Grisafi}, \citenamefont
  {Nigam}, \citenamefont {Ceriotti}, \citenamefont {Schütt}, \citenamefont
  {Westermayr}, \citenamefont {Gastegger}, \citenamefont {Maurer},
  \citenamefont {Kalita}, \citenamefont {Burke}, \citenamefont {Nagai},
  \citenamefont {Akashi}, \citenamefont {Sugino}, \citenamefont {Hermann},
  \citenamefont {Noé}, \citenamefont {Pilati}, \citenamefont {Draxl},
  \citenamefont {Kuban}, \citenamefont {Rigamonti}, \citenamefont {Scheidgen},
  \citenamefont {Esters}, \citenamefont {Hicks}, \citenamefont {Toher},
  \citenamefont {Balachandran}, \citenamefont {Tamblyn}, \citenamefont
  {Whitelam}, \citenamefont {Bellinger},\ and\ \citenamefont
  {Ghiringhelli}}]{kulik_roadmap_2022}%
  \BibitemOpen
  \bibfield  {author} {\bibinfo {author} {\bibfnamefont {H.}~\bibnamefont
  {Kulik}}, \bibinfo {author} {\bibfnamefont {T.}~\bibnamefont
  {Hammerschmidt}}, \bibinfo {author} {\bibfnamefont {J.}~\bibnamefont
  {Schmidt}}, \bibinfo {author} {\bibfnamefont {S.}~\bibnamefont {Botti}},
  \bibinfo {author} {\bibfnamefont {M.~A.~L.}\ \bibnamefont {Marques}},
  \bibinfo {author} {\bibfnamefont {M.}~\bibnamefont {Boley}}, \bibinfo
  {author} {\bibfnamefont {M.}~\bibnamefont {Scheffler}}, \bibinfo {author}
  {\bibfnamefont {M.}~\bibnamefont {Todorović}}, \bibinfo {author}
  {\bibfnamefont {P.}~\bibnamefont {Rinke}}, \bibinfo {author} {\bibfnamefont
  {C.}~\bibnamefont {Oses}}, \bibinfo {author} {\bibfnamefont {A.}~\bibnamefont
  {Smolyanyuk}}, \bibinfo {author} {\bibfnamefont {S.}~\bibnamefont
  {Curtarolo}}, \bibinfo {author} {\bibfnamefont {A.}~\bibnamefont
  {Tkatchenko}}, \bibinfo {author} {\bibfnamefont {A.}~\bibnamefont {Bartok}},
  \bibinfo {author} {\bibfnamefont {S.}~\bibnamefont {Manzhos}}, \bibinfo
  {author} {\bibfnamefont {M.}~\bibnamefont {Ihara}}, \bibinfo {author}
  {\bibfnamefont {T.}~\bibnamefont {Carrington}}, \bibinfo {author}
  {\bibfnamefont {J.}~\bibnamefont {Behler}}, \bibinfo {author} {\bibfnamefont
  {O.}~\bibnamefont {Isayev}}, \bibinfo {author} {\bibfnamefont
  {M.}~\bibnamefont {Veit}}, \bibinfo {author} {\bibfnamefont {A.}~\bibnamefont
  {Grisafi}}, \bibinfo {author} {\bibfnamefont {J.}~\bibnamefont {Nigam}},
  \bibinfo {author} {\bibfnamefont {M.}~\bibnamefont {Ceriotti}}, \bibinfo
  {author} {\bibfnamefont {K.~T.}\ \bibnamefont {Schütt}}, \bibinfo {author}
  {\bibfnamefont {J.}~\bibnamefont {Westermayr}}, \bibinfo {author}
  {\bibfnamefont {M.}~\bibnamefont {Gastegger}}, \bibinfo {author}
  {\bibfnamefont {R.}~\bibnamefont {Maurer}}, \bibinfo {author} {\bibfnamefont
  {B.}~\bibnamefont {Kalita}}, \bibinfo {author} {\bibfnamefont
  {K.}~\bibnamefont {Burke}}, \bibinfo {author} {\bibfnamefont
  {R.}~\bibnamefont {Nagai}}, \bibinfo {author} {\bibfnamefont
  {R.}~\bibnamefont {Akashi}}, \bibinfo {author} {\bibfnamefont
  {O.}~\bibnamefont {Sugino}}, \bibinfo {author} {\bibfnamefont
  {J.}~\bibnamefont {Hermann}}, \bibinfo {author} {\bibfnamefont
  {F.}~\bibnamefont {Noé}}, \bibinfo {author} {\bibfnamefont {S.}~\bibnamefont
  {Pilati}}, \bibinfo {author} {\bibfnamefont {C.}~\bibnamefont {Draxl}},
  \bibinfo {author} {\bibfnamefont {M.}~\bibnamefont {Kuban}}, \bibinfo
  {author} {\bibfnamefont {S.}~\bibnamefont {Rigamonti}}, \bibinfo {author}
  {\bibfnamefont {M.}~\bibnamefont {Scheidgen}}, \bibinfo {author}
  {\bibfnamefont {M.}~\bibnamefont {Esters}}, \bibinfo {author} {\bibfnamefont
  {D.}~\bibnamefont {Hicks}}, \bibinfo {author} {\bibfnamefont
  {C.}~\bibnamefont {Toher}}, \bibinfo {author} {\bibfnamefont
  {P.}~\bibnamefont {Balachandran}}, \bibinfo {author} {\bibfnamefont
  {I.}~\bibnamefont {Tamblyn}}, \bibinfo {author} {\bibfnamefont
  {S.}~\bibnamefont {Whitelam}}, \bibinfo {author} {\bibfnamefont
  {C.}~\bibnamefont {Bellinger}},\ and\ \bibinfo {author} {\bibfnamefont
  {L.~M.}\ \bibnamefont {Ghiringhelli}},\ }\href
  {http://iopscience.iop.org/article/10.1088/2516-1075/ac572f} {\bibfield
  {journal} {\bibinfo  {journal} {Electron. Struct.}\ }\textbf {\bibinfo
  {volume} {4}},\ \bibinfo {pages} {023004} (\bibinfo {year}
  {2022})}\BibitemShut {NoStop}%
\bibitem [{\citenamefont {Himanen}\ \emph {et~al.}(2019)\citenamefont
  {Himanen}, \citenamefont {Geurts}, \citenamefont {Foster},\ and\
  \citenamefont {Rinke}}]{himanen_datadrivenmatSci_2019}%
  \BibitemOpen
  \bibfield  {author} {\bibinfo {author} {\bibfnamefont {L.}~\bibnamefont
  {Himanen}}, \bibinfo {author} {\bibfnamefont {A.}~\bibnamefont {Geurts}},
  \bibinfo {author} {\bibfnamefont {A.~S.}\ \bibnamefont {Foster}},\ and\
  \bibinfo {author} {\bibfnamefont {P.}~\bibnamefont {Rinke}},\ }\href@noop {}
  {\bibfield  {journal} {\bibinfo  {journal} {Adv. Sci.}\ }\textbf {\bibinfo
  {volume} {6}},\ \bibinfo {pages} {1900808} (\bibinfo {year}
  {2019})}\BibitemShut {NoStop}%
\bibitem [{\citenamefont {Bart{\'o}k}\ \emph {et~al.}(2017)\citenamefont
  {Bart{\'o}k}, \citenamefont {De}, \citenamefont {Poelking}, \citenamefont
  {Bernstein}, \citenamefont {Kermode}, \citenamefont {Cs{\'a}nyi},\ and\
  \citenamefont {Ceriotti}}]{bartok_machine_2017}%
  \BibitemOpen
  \bibfield  {author} {\bibinfo {author} {\bibfnamefont {A.~P.}\ \bibnamefont
  {Bart{\'o}k}}, \bibinfo {author} {\bibfnamefont {S.}~\bibnamefont {De}},
  \bibinfo {author} {\bibfnamefont {C.}~\bibnamefont {Poelking}}, \bibinfo
  {author} {\bibfnamefont {N.}~\bibnamefont {Bernstein}}, \bibinfo {author}
  {\bibfnamefont {J.~R.}\ \bibnamefont {Kermode}}, \bibinfo {author}
  {\bibfnamefont {G.}~\bibnamefont {Cs{\'a}nyi}},\ and\ \bibinfo {author}
  {\bibfnamefont {M.}~\bibnamefont {Ceriotti}},\ }\href@noop {} {\bibfield
  {journal} {\bibinfo  {journal} {Sci. Adv.}\ }\textbf {\bibinfo {volume}
  {3}},\ \bibinfo {pages} {e1701816} (\bibinfo {year} {2017})}\BibitemShut
  {NoStop}%
\bibitem [{\citenamefont {Domingos}(2012)}]{domingos_cacm_2012}%
  \BibitemOpen
  \bibfield  {author} {\bibinfo {author} {\bibfnamefont {P.}~\bibnamefont
  {Domingos}},\ }\href {https://doi.org/10.1145/2347736.2347755} {\bibfield
  {journal} {\bibinfo  {journal} {Commun. ACM}\ }\textbf {\bibinfo {volume}
  {55}},\ \bibinfo {pages} {78–87} (\bibinfo {year} {2012})}\BibitemShut
  {NoStop}%
\bibitem [{\citenamefont {Sch\"{u}tt}\ \emph {et~al.}(2017)\citenamefont
  {Sch\"{u}tt}, \citenamefont {Kindermans}, \citenamefont {Sauceda},
  \citenamefont {Chmiela}, \citenamefont {Tkatchenko},\ and\ \citenamefont
  {M\"{u}ller}}]{schutt_schnet_2017}%
  \BibitemOpen
  \bibfield  {author} {\bibinfo {author} {\bibfnamefont {K.~T.}\ \bibnamefont
  {Sch\"{u}tt}}, \bibinfo {author} {\bibfnamefont {P.-J.}\ \bibnamefont
  {Kindermans}}, \bibinfo {author} {\bibfnamefont {H.~E.}\ \bibnamefont
  {Sauceda}}, \bibinfo {author} {\bibfnamefont {S.}~\bibnamefont {Chmiela}},
  \bibinfo {author} {\bibfnamefont {A.}~\bibnamefont {Tkatchenko}},\ and\
  \bibinfo {author} {\bibfnamefont {K.-R.}\ \bibnamefont {M\"{u}ller}},\
  }\href@noop {} {\bibfield  {journal} {\bibinfo  {journal} {NIPS}\ }\textbf
  {\bibinfo {volume} {31}},\ \bibinfo {pages} {992–1002} (\bibinfo {year}
  {2017})}\BibitemShut {NoStop}%
\bibitem [{\citenamefont {Ghosh}\ \emph {et~al.}(2019)\citenamefont {Ghosh},
  \citenamefont {Stuke}, \citenamefont {Todorović}, \citenamefont
  {Jørgensen}, \citenamefont {Schmidt}, \citenamefont {Vehtari},\ and\
  \citenamefont {Rinke}}]{ghosh_deep_2019}%
  \BibitemOpen
  \bibfield  {author} {\bibinfo {author} {\bibfnamefont {K.}~\bibnamefont
  {Ghosh}}, \bibinfo {author} {\bibfnamefont {A.}~\bibnamefont {Stuke}},
  \bibinfo {author} {\bibfnamefont {M.}~\bibnamefont {Todorović}}, \bibinfo
  {author} {\bibfnamefont {P.~B.}\ \bibnamefont {Jørgensen}}, \bibinfo
  {author} {\bibfnamefont {M.~N.}\ \bibnamefont {Schmidt}}, \bibinfo {author}
  {\bibfnamefont {A.}~\bibnamefont {Vehtari}},\ and\ \bibinfo {author}
  {\bibfnamefont {P.}~\bibnamefont {Rinke}},\ }\href
  {https://doi.org/10.1002/advs.201801367} {\bibfield  {journal} {\bibinfo
  {journal} {Adv. Sci.}\ }\textbf {\bibinfo {volume} {6}},\ \bibinfo {pages}
  {1801367} (\bibinfo {year} {2019})}\BibitemShut {NoStop}%
\bibitem [{\citenamefont {Westermayr}\ and\ \citenamefont
  {J.Maurer}(2021)}]{westermayr_physically_2021}%
  \BibitemOpen
  \bibfield  {author} {\bibinfo {author} {\bibfnamefont {J.}~\bibnamefont
  {Westermayr}}\ and\ \bibinfo {author} {\bibfnamefont {R.}~\bibnamefont
  {J.Maurer}},\ }\href@noop {} {\bibfield  {journal} {\bibinfo  {journal}
  {Chem. Sci.}\ }\textbf {\bibinfo {volume} {12}},\ \bibinfo {pages} {10755}
  (\bibinfo {year} {2021})}\BibitemShut {NoStop}%
\bibitem [{\citenamefont {Atz}\ \emph {et~al.}(2021)\citenamefont {Atz},
  \citenamefont {Grisoni},\ and\ \citenamefont
  {Schneider}}]{atz_geometric_2021}%
  \BibitemOpen
  \bibfield  {author} {\bibinfo {author} {\bibfnamefont {K.}~\bibnamefont
  {Atz}}, \bibinfo {author} {\bibfnamefont {F.}~\bibnamefont {Grisoni}},\ and\
  \bibinfo {author} {\bibfnamefont {G.}~\bibnamefont {Schneider}},\ }\href
  {https://doi.org/10.1038/s42256-021-00418-8} {\bibfield  {journal} {\bibinfo
  {journal} {Nat. Mach. Intell.}\ }\textbf {\bibinfo {volume} {3}},\ \bibinfo
  {pages} {1023} (\bibinfo {year} {2021})}\BibitemShut {NoStop}%
\bibitem [{\citenamefont {Behler}(2015)}]{behler_constructing_2015}%
  \BibitemOpen
  \bibfield  {author} {\bibinfo {author} {\bibfnamefont {J.}~\bibnamefont
  {Behler}},\ }\href {https://doi.org/10.1002/qua.24890} {\bibfield  {journal}
  {\bibinfo  {journal} {Int. J. Quantum Chem.}\ }\textbf {\bibinfo {volume}
  {115}},\ \bibinfo {pages} {1032} (\bibinfo {year} {2015})}\BibitemShut
  {NoStop}%
\bibitem [{\citenamefont {Westermayr}\ \emph {et~al.}(2023)\citenamefont
  {Westermayr}, \citenamefont {Gilkes}, \citenamefont {Barrett},\ and\
  \citenamefont {Maurer}}]{westermayr_high-throughput_2023}%
  \BibitemOpen
  \bibfield  {author} {\bibinfo {author} {\bibfnamefont {J.}~\bibnamefont
  {Westermayr}}, \bibinfo {author} {\bibfnamefont {J.}~\bibnamefont {Gilkes}},
  \bibinfo {author} {\bibfnamefont {R.}~\bibnamefont {Barrett}},\ and\ \bibinfo
  {author} {\bibfnamefont {R.~J.}\ \bibnamefont {Maurer}},\ }\href@noop {}
  {\bibfield  {journal} {\bibinfo  {journal} {Nat. Comput. Sci.}\ }\textbf
  {\bibinfo {volume} {3}},\ \bibinfo {pages} {139} (\bibinfo {year}
  {2023})}\BibitemShut {NoStop}%
\bibitem [{\citenamefont {Ramakrishnan}\ \emph {et~al.}(2014)\citenamefont
  {Ramakrishnan}, \citenamefont {Dral}, \citenamefont {Rupp},\ and\
  \citenamefont {von Lilienfeld}}]{ramakrishnan_quantum_2014}%
  \BibitemOpen
  \bibfield  {author} {\bibinfo {author} {\bibfnamefont {R.}~\bibnamefont
  {Ramakrishnan}}, \bibinfo {author} {\bibfnamefont {P.~O.}\ \bibnamefont
  {Dral}}, \bibinfo {author} {\bibfnamefont {M.}~\bibnamefont {Rupp}},\ and\
  \bibinfo {author} {\bibfnamefont {O.~A.}\ \bibnamefont {von Lilienfeld}},\
  }\href@noop {} {\bibfield  {journal} {\bibinfo  {journal} {Sci. Data}\
  }\textbf {\bibinfo {volume} {1}},\ \bibinfo {pages} {201422} (\bibinfo {year}
  {2014})}\BibitemShut {NoStop}%
\bibitem [{\citenamefont {Stuke}\ \emph {et~al.}(2019)\citenamefont {Stuke},
  \citenamefont {Todorović}, \citenamefont {Rupp}, \citenamefont {Kunkel},
  \citenamefont {Ghosh}, \citenamefont {Himanen},\ and\ \citenamefont
  {Rinke}}]{stuke_chemical_2019}%
  \BibitemOpen
  \bibfield  {author} {\bibinfo {author} {\bibfnamefont {A.}~\bibnamefont
  {Stuke}}, \bibinfo {author} {\bibfnamefont {M.}~\bibnamefont {Todorović}},
  \bibinfo {author} {\bibfnamefont {M.}~\bibnamefont {Rupp}}, \bibinfo {author}
  {\bibfnamefont {C.}~\bibnamefont {Kunkel}}, \bibinfo {author} {\bibfnamefont
  {K.}~\bibnamefont {Ghosh}}, \bibinfo {author} {\bibfnamefont
  {L.}~\bibnamefont {Himanen}},\ and\ \bibinfo {author} {\bibfnamefont
  {P.}~\bibnamefont {Rinke}},\ }\href {https://doi.org/10.1063/1.5086105}
  {\bibfield  {journal} {\bibinfo  {journal} {J. Chem. Phys.}\ }\textbf
  {\bibinfo {volume} {150}},\ \bibinfo {pages} {204121} (\bibinfo {year}
  {2019})}\BibitemShut {NoStop}%
\bibitem [{\citenamefont {Fujiwara}\ \emph {et~al.}(2008)\citenamefont
  {Fujiwara}, \citenamefont {Yamashita}, \citenamefont {Osoda}, \citenamefont
  {Asogawa}, \citenamefont {Fukushima}, \citenamefont {Asao}, \citenamefont
  {Shimadzu}, \citenamefont {Nakao},\ and\ \citenamefont
  {Shimizu}}]{fujiwara_jcim_2008}%
  \BibitemOpen
  \bibfield  {author} {\bibinfo {author} {\bibfnamefont {Y.}~\bibnamefont
  {Fujiwara}}, \bibinfo {author} {\bibfnamefont {Y.}~\bibnamefont {Yamashita}},
  \bibinfo {author} {\bibfnamefont {T.}~\bibnamefont {Osoda}}, \bibinfo
  {author} {\bibfnamefont {M.}~\bibnamefont {Asogawa}}, \bibinfo {author}
  {\bibfnamefont {C.}~\bibnamefont {Fukushima}}, \bibinfo {author}
  {\bibfnamefont {M.}~\bibnamefont {Asao}}, \bibinfo {author} {\bibfnamefont
  {H.}~\bibnamefont {Shimadzu}}, \bibinfo {author} {\bibfnamefont
  {K.}~\bibnamefont {Nakao}},\ and\ \bibinfo {author} {\bibfnamefont
  {R.}~\bibnamefont {Shimizu}},\ }\href {https://doi.org/10.1021/ci700085q}
  {\bibfield  {journal} {\bibinfo  {journal} {J. Chem. Inf. Model.}\ }\textbf
  {\bibinfo {volume} {48}},\ \bibinfo {pages} {930–940} (\bibinfo {year}
  {2008})}\BibitemShut {NoStop}%
\bibitem [{\citenamefont {Reker}\ \emph {et~al.}(2016)\citenamefont {Reker},
  \citenamefont {Schneider},\ and\ \citenamefont
  {Schneider}}]{reker_multi-objective_2016}%
  \BibitemOpen
  \bibfield  {author} {\bibinfo {author} {\bibfnamefont {D.}~\bibnamefont
  {Reker}}, \bibinfo {author} {\bibfnamefont {P.}~\bibnamefont {Schneider}},\
  and\ \bibinfo {author} {\bibfnamefont {G.}~\bibnamefont {Schneider}},\ }\href
  {https://doi.org/10.1039/C5SC04272K} {\bibfield  {journal} {\bibinfo
  {journal} {Chem. Sci.}\ }\textbf {\bibinfo {volume} {7}},\ \bibinfo {pages}
  {3919} (\bibinfo {year} {2016})}\BibitemShut {NoStop}%
\bibitem [{\citenamefont {Zhou}\ \emph {et~al.}(2022)\citenamefont {Zhou},
  \citenamefont {Wang}, \citenamefont {Tang}, \citenamefont {Feng},
  \citenamefont {Hooi}, \citenamefont {Zhao}, \citenamefont {Xu},\ and\
  \citenamefont {Wang}}]{zhou_jointly_2022}%
  \BibitemOpen
  \bibfield  {author} {\bibinfo {author} {\bibfnamefont {K.}~\bibnamefont
  {Zhou}}, \bibinfo {author} {\bibfnamefont {K.}~\bibnamefont {Wang}}, \bibinfo
  {author} {\bibfnamefont {J.}~\bibnamefont {Tang}}, \bibinfo {author}
  {\bibfnamefont {J.}~\bibnamefont {Feng}}, \bibinfo {author} {\bibfnamefont
  {B.}~\bibnamefont {Hooi}}, \bibinfo {author} {\bibfnamefont {P.}~\bibnamefont
  {Zhao}}, \bibinfo {author} {\bibfnamefont {T.}~\bibnamefont {Xu}},\ and\
  \bibinfo {author} {\bibfnamefont {X.}~\bibnamefont {Wang}},\ }\href@noop {}
  {\bibfield  {journal} {\bibinfo  {journal} {LoG}\ }\textbf {\bibinfo {volume}
  {198}},\ \bibinfo {pages} {29} (\bibinfo {year} {2022})}\BibitemShut
  {NoStop}%
\bibitem [{\citenamefont {Hwang}\ \emph {et~al.}(2022)\citenamefont {Hwang},
  \citenamefont {Choi},\ and\ \citenamefont
  {Choi}}]{hwang_uncertainty-based_2022}%
  \BibitemOpen
  \bibfield  {author} {\bibinfo {author} {\bibfnamefont {S.}~\bibnamefont
  {Hwang}}, \bibinfo {author} {\bibfnamefont {J.}~\bibnamefont {Choi}},\ and\
  \bibinfo {author} {\bibfnamefont {J.}~\bibnamefont {Choi}},\ }\href@noop {}
  {\bibfield  {journal} {\bibinfo  {journal} {{IEEE} Access}\ }\textbf
  {\bibinfo {volume} {10}},\ \bibinfo {pages} {110983} (\bibinfo {year}
  {2022})}\BibitemShut {NoStop}%
\bibitem [{\citenamefont {Desai}\ \emph {et~al.}(2013)\citenamefont {Desai},
  \citenamefont {Dixon}, \citenamefont {Farrant}, \citenamefont {Feng},
  \citenamefont {Gibson}, \citenamefont {van Hoorn}, \citenamefont {Mills},
  \citenamefont {Morgan}, \citenamefont {Parry}, \citenamefont {Ramjee},
  \citenamefont {Selway}, \citenamefont {Tarver}, \citenamefont {Whitlock},\
  and\ \citenamefont {Wright}}]{desai_rapid_2013}%
  \BibitemOpen
  \bibfield  {author} {\bibinfo {author} {\bibfnamefont {B.}~\bibnamefont
  {Desai}}, \bibinfo {author} {\bibfnamefont {K.}~\bibnamefont {Dixon}},
  \bibinfo {author} {\bibfnamefont {E.}~\bibnamefont {Farrant}}, \bibinfo
  {author} {\bibfnamefont {Q.}~\bibnamefont {Feng}}, \bibinfo {author}
  {\bibfnamefont {K.~R.}\ \bibnamefont {Gibson}}, \bibinfo {author}
  {\bibfnamefont {W.~P.}\ \bibnamefont {van Hoorn}}, \bibinfo {author}
  {\bibfnamefont {J.}~\bibnamefont {Mills}}, \bibinfo {author} {\bibfnamefont
  {T.}~\bibnamefont {Morgan}}, \bibinfo {author} {\bibfnamefont {D.~M.}\
  \bibnamefont {Parry}}, \bibinfo {author} {\bibfnamefont {M.~K.}\ \bibnamefont
  {Ramjee}}, \bibinfo {author} {\bibfnamefont {C.~N.}\ \bibnamefont {Selway}},
  \bibinfo {author} {\bibfnamefont {G.~J.}\ \bibnamefont {Tarver}}, \bibinfo
  {author} {\bibfnamefont {G.}~\bibnamefont {Whitlock}},\ and\ \bibinfo
  {author} {\bibfnamefont {A.~G.}\ \bibnamefont {Wright}},\ }\href
  {https://doi.org/10.1021/jm400099d} {\bibfield  {journal} {\bibinfo
  {journal} {J. Med. Chem.}\ }\textbf {\bibinfo {volume} {56}},\ \bibinfo
  {pages} {3033} (\bibinfo {year} {2013})}\BibitemShut {NoStop}%
\bibitem [{\citenamefont {Besnard}\ \emph {et~al.}(2012)\citenamefont
  {Besnard}, \citenamefont {Ruda}, \citenamefont {Setola}, \citenamefont
  {Abecassis}, \citenamefont {Rodriguiz}, \citenamefont {Huang}, \citenamefont
  {Norval}, \citenamefont {Sassano}, \citenamefont {Shin}, \citenamefont
  {Webster}, \citenamefont {Simeons}, \citenamefont {Stojanovski},
  \citenamefont {Prat}, \citenamefont {Seidah}, \citenamefont {Constam},
  \citenamefont {Bickerton}, \citenamefont {Read}, \citenamefont {Wetsel},
  \citenamefont {Gilbert}, \citenamefont {Roth},\ and\ \citenamefont
  {Hopkins}}]{besnard_automated_2012}%
  \BibitemOpen
  \bibfield  {author} {\bibinfo {author} {\bibfnamefont {J.}~\bibnamefont
  {Besnard}}, \bibinfo {author} {\bibfnamefont {G.~F.}\ \bibnamefont {Ruda}},
  \bibinfo {author} {\bibfnamefont {V.}~\bibnamefont {Setola}}, \bibinfo
  {author} {\bibfnamefont {K.}~\bibnamefont {Abecassis}}, \bibinfo {author}
  {\bibfnamefont {R.~M.}\ \bibnamefont {Rodriguiz}}, \bibinfo {author}
  {\bibfnamefont {X.-P.}\ \bibnamefont {Huang}}, \bibinfo {author}
  {\bibfnamefont {S.}~\bibnamefont {Norval}}, \bibinfo {author} {\bibfnamefont
  {M.~F.}\ \bibnamefont {Sassano}}, \bibinfo {author} {\bibfnamefont {A.~I.}\
  \bibnamefont {Shin}}, \bibinfo {author} {\bibfnamefont {L.~A.}\ \bibnamefont
  {Webster}}, \bibinfo {author} {\bibfnamefont {F.~R.}\ \bibnamefont
  {Simeons}}, \bibinfo {author} {\bibfnamefont {L.}~\bibnamefont
  {Stojanovski}}, \bibinfo {author} {\bibfnamefont {A.}~\bibnamefont {Prat}},
  \bibinfo {author} {\bibfnamefont {N.~G.}\ \bibnamefont {Seidah}}, \bibinfo
  {author} {\bibfnamefont {D.~B.}\ \bibnamefont {Constam}}, \bibinfo {author}
  {\bibfnamefont {G.~R.}\ \bibnamefont {Bickerton}}, \bibinfo {author}
  {\bibfnamefont {K.~D.}\ \bibnamefont {Read}}, \bibinfo {author}
  {\bibfnamefont {W.~C.}\ \bibnamefont {Wetsel}}, \bibinfo {author}
  {\bibfnamefont {I.~H.}\ \bibnamefont {Gilbert}}, \bibinfo {author}
  {\bibfnamefont {B.~L.}\ \bibnamefont {Roth}},\ and\ \bibinfo {author}
  {\bibfnamefont {A.~L.}\ \bibnamefont {Hopkins}},\ }\href@noop {} {\bibfield
  {journal} {\bibinfo  {journal} {Nature}\ }\textbf {\bibinfo {volume} {492}},\
  \bibinfo {pages} {215} (\bibinfo {year} {2012})}\BibitemShut {NoStop}%
\bibitem [{\citenamefont {Naik}\ \emph {et~al.}(2016)\citenamefont {Naik},
  \citenamefont {Kangas}, \citenamefont {Sullivan},\ and\ \citenamefont
  {Murphy}}]{naik_active_2016}%
  \BibitemOpen
  \bibfield  {author} {\bibinfo {author} {\bibfnamefont {A.~W.}\ \bibnamefont
  {Naik}}, \bibinfo {author} {\bibfnamefont {J.~D.}\ \bibnamefont {Kangas}},
  \bibinfo {author} {\bibfnamefont {D.~P.}\ \bibnamefont {Sullivan}},\ and\
  \bibinfo {author} {\bibfnamefont {R.~F.}\ \bibnamefont {Murphy}},\
  }\href@noop {} {\bibfield  {journal} {\bibinfo  {journal} {{eLife}}\ }\textbf
  {\bibinfo {volume} {10047}} (\bibinfo {year} {2016})}\BibitemShut {NoStop}%
\bibitem [{\citenamefont {Viet~Johansson}\ \emph {et~al.}(2022)\citenamefont
  {Viet~Johansson}, \citenamefont {Gummesson~Svensson}, \citenamefont
  {Bjerrum}, \citenamefont {Schliep}, \citenamefont {Haghir~Chehreghani},
  \citenamefont {Tyrchan},\ and\ \citenamefont
  {Engkvist}}]{viet_johansson_using_2022}%
  \BibitemOpen
  \bibfield  {author} {\bibinfo {author} {\bibfnamefont {S.}~\bibnamefont
  {Viet~Johansson}}, \bibinfo {author} {\bibfnamefont {H.}~\bibnamefont
  {Gummesson~Svensson}}, \bibinfo {author} {\bibfnamefont {E.}~\bibnamefont
  {Bjerrum}}, \bibinfo {author} {\bibfnamefont {A.}~\bibnamefont {Schliep}},
  \bibinfo {author} {\bibfnamefont {M.}~\bibnamefont {Haghir~Chehreghani}},
  \bibinfo {author} {\bibfnamefont {C.}~\bibnamefont {Tyrchan}},\ and\ \bibinfo
  {author} {\bibfnamefont {O.}~\bibnamefont {Engkvist}},\ }\href@noop {}
  {\bibfield  {journal} {\bibinfo  {journal} {Mol. Inf.}\ }\textbf {\bibinfo
  {volume} {41}},\ \bibinfo {pages} {2200043} (\bibinfo {year}
  {2022})}\BibitemShut {NoStop}%
\bibitem [{\citenamefont {Vandenhaute}\ \emph {et~al.}(2023)\citenamefont
  {Vandenhaute}, \citenamefont {Cools-Ceuppens}, \citenamefont {DeKeyser},
  \citenamefont {Verstraelen},\ and\ \citenamefont
  {Van~Speybroeck}}]{vandenhaute_machine_2023}%
  \BibitemOpen
  \bibfield  {author} {\bibinfo {author} {\bibfnamefont {S.}~\bibnamefont
  {Vandenhaute}}, \bibinfo {author} {\bibfnamefont {M.}~\bibnamefont
  {Cools-Ceuppens}}, \bibinfo {author} {\bibfnamefont {S.}~\bibnamefont
  {DeKeyser}}, \bibinfo {author} {\bibfnamefont {T.}~\bibnamefont
  {Verstraelen}},\ and\ \bibinfo {author} {\bibfnamefont {V.}~\bibnamefont
  {Van~Speybroeck}},\ }\href {https://doi.org/10.1038/s41524-023-00969-x}
  {\bibfield  {journal} {\bibinfo  {journal} {npj Comput. Mater.}\ }\textbf
  {\bibinfo {volume} {9}},\ \bibinfo {pages} {1} (\bibinfo {year}
  {2023})}\BibitemShut {NoStop}%
\bibitem [{\citenamefont {Wen}\ \emph {et~al.}(2023)\citenamefont {Wen},
  \citenamefont {Li}, \citenamefont {Xiang},\ and\ \citenamefont
  {Reker}}]{wen_2023}%
  \BibitemOpen
  \bibfield  {author} {\bibinfo {author} {\bibfnamefont {Y.}~\bibnamefont
  {Wen}}, \bibinfo {author} {\bibfnamefont {Z.}~\bibnamefont {Li}}, \bibinfo
  {author} {\bibfnamefont {Y.}~\bibnamefont {Xiang}},\ and\ \bibinfo {author}
  {\bibfnamefont {D.}~\bibnamefont {Reker}},\ }\href@noop {} {\bibfield
  {journal} {\bibinfo  {journal} {Digital Discov.}\ }\textbf {\bibinfo {volume}
  {4}},\ \bibinfo {pages} {1134} (\bibinfo {year} {2023})}\BibitemShut
  {NoStop}%
\bibitem [{\citenamefont {Jose}\ \emph {et~al.}(2023)\citenamefont {Jose},
  \citenamefont {de~Mendon\c{c}a}, \citenamefont {Devijver}, \citenamefont
  {Jakse}, \citenamefont {Monbet},\ and\ \citenamefont {Poloni}}]{jose_2023}%
  \BibitemOpen
  \bibfield  {author} {\bibinfo {author} {\bibfnamefont {A.}~\bibnamefont
  {Jose}}, \bibinfo {author} {\bibfnamefont {J.~P.~A.}\ \bibnamefont
  {de~Mendon\c{c}a}}, \bibinfo {author} {\bibfnamefont {E.}~\bibnamefont
  {Devijver}}, \bibinfo {author} {\bibfnamefont {N.}~\bibnamefont {Jakse}},
  \bibinfo {author} {\bibfnamefont {V.}~\bibnamefont {Monbet}},\ and\ \bibinfo
  {author} {\bibfnamefont {R.}~\bibnamefont {Poloni}},\ }\href@noop {}
  {\bibfield  {journal} {\bibinfo  {journal} {Data Min. Knowl. Discov.}\
  }\textbf {\bibinfo {volume} {38}},\ \bibinfo {pages} {420–460} (\bibinfo
  {year} {2023})}\BibitemShut {NoStop}%
\bibitem [{\citenamefont {Besel}\ \emph {et~al.}(2024)\citenamefont {Besel},
  \citenamefont {Todorović}, \citenamefont {Kurtén}, \citenamefont
  {Vehkamäki},\ and\ \citenamefont {Rinke}}]{besel_2024}%
  \BibitemOpen
  \bibfield  {author} {\bibinfo {author} {\bibfnamefont {V.}~\bibnamefont
  {Besel}}, \bibinfo {author} {\bibfnamefont {M.}~\bibnamefont {Todorović}},
  \bibinfo {author} {\bibfnamefont {T.}~\bibnamefont {Kurtén}}, \bibinfo
  {author} {\bibfnamefont {H.}~\bibnamefont {Vehkamäki}},\ and\ \bibinfo
  {author} {\bibfnamefont {P.}~\bibnamefont {Rinke}},\ }\href@noop {}
  {\bibfield  {journal} {\bibinfo  {journal} {J. Aerosol Sci.}\ }\textbf
  {\bibinfo {volume} {179}},\ \bibinfo {pages} {106375} (\bibinfo {year}
  {2024})}\BibitemShut {NoStop}%
\bibitem [{\citenamefont {Zaverkin}\ and\ \citenamefont
  {Kästner}(2021)}]{zaverkin_exploration_2021}%
  \BibitemOpen
  \bibfield  {author} {\bibinfo {author} {\bibfnamefont {V.}~\bibnamefont
  {Zaverkin}}\ and\ \bibinfo {author} {\bibfnamefont {J.}~\bibnamefont
  {Kästner}},\ }\href@noop {} {\bibfield  {journal} {\bibinfo  {journal}
  {Mach. Learn.: Sci. Technol.s}\ }\textbf {\bibinfo {volume} {2}},\ \bibinfo
  {pages} {035009} (\bibinfo {year} {2021})}\BibitemShut {NoStop}%
\bibitem [{\citenamefont {Richards}\ \emph {et~al.}(2011)\citenamefont
  {Richards}, \citenamefont {Starr}, \citenamefont {Brink}, \citenamefont
  {Miller}, \citenamefont {Bloom}, \citenamefont {Butler}, \citenamefont
  {Berian~James}, \citenamefont {Long},\ and\ \citenamefont
  {Rice}}]{richards_2011}%
  \BibitemOpen
  \bibfield  {author} {\bibinfo {author} {\bibfnamefont {J.~W.}\ \bibnamefont
  {Richards}}, \bibinfo {author} {\bibfnamefont {D.~L.}\ \bibnamefont {Starr}},
  \bibinfo {author} {\bibfnamefont {H.}~\bibnamefont {Brink}}, \bibinfo
  {author} {\bibfnamefont {A.~A.}\ \bibnamefont {Miller}}, \bibinfo {author}
  {\bibfnamefont {J.~S.}\ \bibnamefont {Bloom}}, \bibinfo {author}
  {\bibfnamefont {N.~R.}\ \bibnamefont {Butler}}, \bibinfo {author}
  {\bibfnamefont {J.}~\bibnamefont {Berian~James}}, \bibinfo {author}
  {\bibfnamefont {J.~P.}\ \bibnamefont {Long}},\ and\ \bibinfo {author}
  {\bibfnamefont {J.}~\bibnamefont {Rice}},\ }\href@noop {} {\bibfield
  {journal} {\bibinfo  {journal} {ApJ}\ }\textbf {\bibinfo {volume} {744}},\
  \bibinfo {pages} {192} (\bibinfo {year} {2011})}\BibitemShut {NoStop}%
\bibitem [{\citenamefont {Farquhar}\ \emph {et~al.}(2021)\citenamefont
  {Farquhar}, \citenamefont {Gal},\ and\ \citenamefont
  {Rainforth}}]{farquhar_2021}%
  \BibitemOpen
  \bibfield  {author} {\bibinfo {author} {\bibfnamefont {S.}~\bibnamefont
  {Farquhar}}, \bibinfo {author} {\bibfnamefont {Y.}~\bibnamefont {Gal}},\ and\
  \bibinfo {author} {\bibfnamefont {T.}~\bibnamefont {Rainforth}},\ }\href@noop
  {} {\bibfield  {journal} {\bibinfo  {journal} {{ICLR}}\ } (\bibinfo {year}
  {2021})}\BibitemShut {NoStop}%
\bibitem [{\citenamefont {Ropo}\ \emph {et~al.}(2016)\citenamefont {Ropo},
  \citenamefont {Schneider}, \citenamefont {Baldauf},\ and\ \citenamefont
  {Blum}}]{ropo_first-principles_2016}%
  \BibitemOpen
  \bibfield  {author} {\bibinfo {author} {\bibfnamefont {M.}~\bibnamefont
  {Ropo}}, \bibinfo {author} {\bibfnamefont {M.}~\bibnamefont {Schneider}},
  \bibinfo {author} {\bibfnamefont {C.}~\bibnamefont {Baldauf}},\ and\ \bibinfo
  {author} {\bibfnamefont {V.}~\bibnamefont {Blum}},\ }\href
  {https://doi.org/10.1038/sdata.2016.9} {\bibfield  {journal} {\bibinfo
  {journal} {Sci. Data}\ }\textbf {\bibinfo {volume} {3}},\ \bibinfo {pages}
  {160009} (\bibinfo {year} {2016})}\BibitemShut {NoStop}%
\bibitem [{\citenamefont {Perdew}\ \emph {et~al.}(1997)\citenamefont {Perdew},
  \citenamefont {Burke},\ and\ \citenamefont
  {Ernzerhof}}]{perdew_generalized_1997}%
  \BibitemOpen
  \bibfield  {author} {\bibinfo {author} {\bibfnamefont {J.~P.}\ \bibnamefont
  {Perdew}}, \bibinfo {author} {\bibfnamefont {K.}~\bibnamefont {Burke}},\ and\
  \bibinfo {author} {\bibfnamefont {M.}~\bibnamefont {Ernzerhof}},\ }\href
  {https://doi.org/10.1103/PhysRevLett.78.1396} {\bibfield  {journal} {\bibinfo
   {journal} {Phys. Rev. Lett.}\ }\textbf {\bibinfo {volume} {78}},\ \bibinfo
  {pages} {1396} (\bibinfo {year} {1997})}\BibitemShut {NoStop}%
\bibitem [{\citenamefont {Tkatchenko}\ and\ \citenamefont
  {Scheffler}(2009)}]{tkatchenko_accurate_2009}%
  \BibitemOpen
  \bibfield  {author} {\bibinfo {author} {\bibfnamefont {A.}~\bibnamefont
  {Tkatchenko}}\ and\ \bibinfo {author} {\bibfnamefont {M.}~\bibnamefont
  {Scheffler}},\ }\href {https://doi.org/10.1103/PhysRevLett.102.073005}
  {\bibfield  {journal} {\bibinfo  {journal} {Phys. Rev. Lett.}\ }\textbf
  {\bibinfo {volume} {102}},\ \bibinfo {pages} {073005} (\bibinfo {year}
  {2009})}\BibitemShut {NoStop}%
\bibitem [{\citenamefont {De}\ \emph {et~al.}(2016)\citenamefont {De},
  \citenamefont {Bart{\'o}k}, \citenamefont {Cs{\'a}nyi},\ and\ \citenamefont
  {Ceriotti}}]{de_comparing_2016}%
  \BibitemOpen
  \bibfield  {author} {\bibinfo {author} {\bibfnamefont {S.}~\bibnamefont
  {De}}, \bibinfo {author} {\bibfnamefont {A.~P.}\ \bibnamefont {Bart{\'o}k}},
  \bibinfo {author} {\bibfnamefont {G.}~\bibnamefont {Cs{\'a}nyi}},\ and\
  \bibinfo {author} {\bibfnamefont {M.}~\bibnamefont {Ceriotti}},\ }\href@noop
  {} {\bibfield  {journal} {\bibinfo  {journal} {Phys. Chem. Chem. Phys.}\
  }\textbf {\bibinfo {volume} {18}},\ \bibinfo {pages} {13754} (\bibinfo {year}
  {2016})}\BibitemShut {NoStop}%
\bibitem [{\citenamefont {Artrith}\ \emph {et~al.}(2017)\citenamefont
  {Artrith}, \citenamefont {Urban},\ and\ \citenamefont
  {Ceder}}]{artrith_efficient_2017}%
  \BibitemOpen
  \bibfield  {author} {\bibinfo {author} {\bibfnamefont {N.}~\bibnamefont
  {Artrith}}, \bibinfo {author} {\bibfnamefont {A.}~\bibnamefont {Urban}},\
  and\ \bibinfo {author} {\bibfnamefont {G.}~\bibnamefont {Ceder}},\ }\href
  {https://doi.org/10.1103/PhysRevB.96.014112} {\bibfield  {journal} {\bibinfo
  {journal} {Phys. Rev. B}\ }\textbf {\bibinfo {volume} {96}},\ \bibinfo
  {pages} {014112} (\bibinfo {year} {2017})}\BibitemShut {NoStop}%
\bibitem [{\citenamefont {De}\ \emph {et~al.}(2017)\citenamefont {De},
  \citenamefont {Musil}, \citenamefont {Ingram}, \citenamefont {Baldauf},\ and\
  \citenamefont {Ceriotti}}]{de_mapping_2017}%
  \BibitemOpen
  \bibfield  {author} {\bibinfo {author} {\bibfnamefont {S.}~\bibnamefont
  {De}}, \bibinfo {author} {\bibfnamefont {F.}~\bibnamefont {Musil}}, \bibinfo
  {author} {\bibfnamefont {T.}~\bibnamefont {Ingram}}, \bibinfo {author}
  {\bibfnamefont {C.}~\bibnamefont {Baldauf}},\ and\ \bibinfo {author}
  {\bibfnamefont {M.}~\bibnamefont {Ceriotti}},\ }\href@noop {} {\bibfield
  {journal} {\bibinfo  {journal} {J. Cheminform.}\ }\textbf {\bibinfo {volume}
  {9}},\ \bibinfo {pages} {6} (\bibinfo {year} {2017})}\BibitemShut {NoStop}%
\bibitem [{\citenamefont {Ruddigkeit}\ \emph {et~al.}(2012)\citenamefont
  {Ruddigkeit}, \citenamefont {van Deursen}, \citenamefont {Blum},\ and\
  \citenamefont {Reymond}}]{ruddigkeit_enumeration_2012}%
  \BibitemOpen
  \bibfield  {author} {\bibinfo {author} {\bibfnamefont {L.}~\bibnamefont
  {Ruddigkeit}}, \bibinfo {author} {\bibfnamefont {R.}~\bibnamefont {van
  Deursen}}, \bibinfo {author} {\bibfnamefont {L.~C.}\ \bibnamefont {Blum}},\
  and\ \bibinfo {author} {\bibfnamefont {J.-L.}\ \bibnamefont {Reymond}},\
  }\href {https://doi.org/10.1021/ci300415d} {\bibfield  {journal} {\bibinfo
  {journal} {J. Chem. Inf. Model.}\ }\textbf {\bibinfo {volume} {52}},\
  \bibinfo {pages} {2864} (\bibinfo {year} {2012})}\BibitemShut {NoStop}%
\bibitem [{\citenamefont {Stuke}\ \emph {et~al.}(2020)\citenamefont {Stuke},
  \citenamefont {Kunkel}, \citenamefont {Golze}, \citenamefont {Todorović},
  \citenamefont {Margraf}, \citenamefont {Reuter}, \citenamefont {Rinke},\ and\
  \citenamefont {Oberhofer}}]{stuke_atomic_2020}%
  \BibitemOpen
  \bibfield  {author} {\bibinfo {author} {\bibfnamefont {A.}~\bibnamefont
  {Stuke}}, \bibinfo {author} {\bibfnamefont {C.}~\bibnamefont {Kunkel}},
  \bibinfo {author} {\bibfnamefont {D.}~\bibnamefont {Golze}}, \bibinfo
  {author} {\bibfnamefont {M.}~\bibnamefont {Todorović}}, \bibinfo {author}
  {\bibfnamefont {J.~T.}\ \bibnamefont {Margraf}}, \bibinfo {author}
  {\bibfnamefont {K.}~\bibnamefont {Reuter}}, \bibinfo {author} {\bibfnamefont
  {P.}~\bibnamefont {Rinke}},\ and\ \bibinfo {author} {\bibfnamefont
  {H.}~\bibnamefont {Oberhofer}},\ }\href
  {https://doi.org/10.1038/s41597-020-0385-y} {\bibfield  {journal} {\bibinfo
  {journal} {Sci. Data}\ }\textbf {\bibinfo {volume} {7}},\ \bibinfo {pages}
  {58} (\bibinfo {year} {2020})}\BibitemShut {NoStop}%
\bibitem [{\citenamefont {Allen}(2002)}]{allen_cambridge_2002}%
  \BibitemOpen
  \bibfield  {author} {\bibinfo {author} {\bibfnamefont {F.~H.}\ \bibnamefont
  {Allen}},\ }\href@noop {} {\bibfield  {journal} {\bibinfo  {journal} {Acta
  Cryst. B}\ }\textbf {\bibinfo {volume} {58}},\ \bibinfo {pages} {380}
  (\bibinfo {year} {2002})}\BibitemShut {NoStop}%
\bibitem [{\citenamefont {Schober}\ \emph {et~al.}(2016)\citenamefont
  {Schober}, \citenamefont {Reuter},\ and\ \citenamefont
  {Oberhofer}}]{schober_virtual_2016}%
  \BibitemOpen
  \bibfield  {author} {\bibinfo {author} {\bibfnamefont {C.}~\bibnamefont
  {Schober}}, \bibinfo {author} {\bibfnamefont {K.}~\bibnamefont {Reuter}},\
  and\ \bibinfo {author} {\bibfnamefont {H.}~\bibnamefont {Oberhofer}},\
  }\href@noop {} {\bibfield  {journal} {\bibinfo  {journal} {J. Phys. Chem.
  Lett.}\ }\textbf {\bibinfo {volume} {7}},\ \bibinfo {pages} {3973} (\bibinfo
  {year} {2016})}\BibitemShut {NoStop}%
\bibitem [{\citenamefont {Schober}(2017)}]{schober_ab_2017}%
  \BibitemOpen
  \bibfield  {author} {\bibinfo {author} {\bibfnamefont {C.}~\bibnamefont
  {Schober}},\ }\href@noop {} {\bibfield  {journal} {\bibinfo  {journal}
  {Dissertation, {TU} {M{\"u}nchen}}\ } (\bibinfo {year} {2017})}\BibitemShut
  {NoStop}%
\bibitem [{\citenamefont {Choi}\ \emph {et~al.}(2022)\citenamefont {Choi},
  \citenamefont {Zhang}, \citenamefont {Mehta}, \citenamefont {Blanchard},\
  and\ \citenamefont {Lupo~Pasini}}]{choi_scalable_2022}%
  \BibitemOpen
  \bibfield  {author} {\bibinfo {author} {\bibfnamefont {J.~Y.}\ \bibnamefont
  {Choi}}, \bibinfo {author} {\bibfnamefont {P.}~\bibnamefont {Zhang}},
  \bibinfo {author} {\bibfnamefont {K.}~\bibnamefont {Mehta}}, \bibinfo
  {author} {\bibfnamefont {A.}~\bibnamefont {Blanchard}},\ and\ \bibinfo
  {author} {\bibfnamefont {M.}~\bibnamefont {Lupo~Pasini}},\ }\href
  {https://doi.org/10.1186/s13321-022-00652-1} {\bibfield  {journal} {\bibinfo
  {journal} {J. Cheminform.}\ }\textbf {\bibinfo {volume} {14}},\ \bibinfo
  {pages} {70} (\bibinfo {year} {2022})}\BibitemShut {NoStop}%
\bibitem [{\citenamefont {Rasmussen}\ and\ \citenamefont
  {Williams}(2006)}]{rasmussen_gpml_2006}%
  \BibitemOpen
  \bibfield  {author} {\bibinfo {author} {\bibfnamefont {C.~E.}\ \bibnamefont
  {Rasmussen}}\ and\ \bibinfo {author} {\bibfnamefont {C.~K.~I.}\ \bibnamefont
  {Williams}},\ }\href {https://www.worldcat.org/oclc/61285753} {\emph
  {\bibinfo {title} {Gaussian processes for machine learning}}}\ (\bibinfo
  {publisher} {{MIT} Press},\ \bibinfo {year} {2006})\BibitemShut {NoStop}%
\bibitem [{\citenamefont {Settles}(2012)}]{settles_active_2009}%
  \BibitemOpen
  \bibfield  {author} {\bibinfo {author} {\bibfnamefont {B.}~\bibnamefont
  {Settles}},\ }\href {https://doi.org/10.1007/978-3-031-01560-1} {\emph
  {\bibinfo {title} {Active Learning}}}\ (\bibinfo  {publisher} {Springer
  International Publishing},\ \bibinfo {year} {2012})\BibitemShut {NoStop}%
\bibitem [{\citenamefont {Wang}\ \emph {et~al.}(2022)\citenamefont {Wang},
  \citenamefont {Liang}, \citenamefont {{McDannald}}, \citenamefont
  {Takeuchi},\ and\ \citenamefont {Kusne}}]{wang_benchmarking_2022}%
  \BibitemOpen
  \bibfield  {author} {\bibinfo {author} {\bibfnamefont {A.}~\bibnamefont
  {Wang}}, \bibinfo {author} {\bibfnamefont {H.}~\bibnamefont {Liang}},
  \bibinfo {author} {\bibfnamefont {A.}~\bibnamefont {{McDannald}}}, \bibinfo
  {author} {\bibfnamefont {I.}~\bibnamefont {Takeuchi}},\ and\ \bibinfo
  {author} {\bibfnamefont {A.~G.}\ \bibnamefont {Kusne}},\ }\href
  {https://doi.org/10.1093/oxfmat/itac006} {\bibfield  {journal} {\bibinfo
  {journal} {Oxford Open Mater. Sci.}\ }\textbf {\bibinfo {volume} {2}},\
  \bibinfo {pages} {itac006} (\bibinfo {year} {2022})}\BibitemShut {NoStop}%
\bibitem [{\citenamefont {Bassman~Oftelie}\ \emph {et~al.}(2018)\citenamefont
  {Bassman~Oftelie}, \citenamefont {Rajak}, \citenamefont {Kalia},
  \citenamefont {Nakano}, \citenamefont {Sha}, \citenamefont {Sun},
  \citenamefont {Singh}, \citenamefont {Aykol}, \citenamefont {Huck},
  \citenamefont {Persson},\ and\ \citenamefont
  {Vashishta}}]{bassman_oftelie_active_2018}%
  \BibitemOpen
  \bibfield  {author} {\bibinfo {author} {\bibfnamefont {L.}~\bibnamefont
  {Bassman~Oftelie}}, \bibinfo {author} {\bibfnamefont {P.}~\bibnamefont
  {Rajak}}, \bibinfo {author} {\bibfnamefont {R.~K.}\ \bibnamefont {Kalia}},
  \bibinfo {author} {\bibfnamefont {A.}~\bibnamefont {Nakano}}, \bibinfo
  {author} {\bibfnamefont {F.}~\bibnamefont {Sha}}, \bibinfo {author}
  {\bibfnamefont {J.}~\bibnamefont {Sun}}, \bibinfo {author} {\bibfnamefont
  {D.~J.}\ \bibnamefont {Singh}}, \bibinfo {author} {\bibfnamefont
  {M.}~\bibnamefont {Aykol}}, \bibinfo {author} {\bibfnamefont
  {P.}~\bibnamefont {Huck}}, \bibinfo {author} {\bibfnamefont {K.}~\bibnamefont
  {Persson}},\ and\ \bibinfo {author} {\bibfnamefont {P.}~\bibnamefont
  {Vashishta}},\ }\href {https://doi.org/10.1038/s41524-018-0129-0} {\bibfield
  {journal} {\bibinfo  {journal} {npj Comput. Mater.}\ }\textbf {\bibinfo
  {volume} {4}},\ \bibinfo {pages} {1} (\bibinfo {year} {2018})}\BibitemShut
  {NoStop}%
\bibitem [{\citenamefont {Reker}(2020)}]{reker_2020}%
  \BibitemOpen
  \bibfield  {author} {\bibinfo {author} {\bibfnamefont {D.}~\bibnamefont
  {Reker}},\ }in\ \href {https://doi.org/10.1039/9781788016841-00301} {\emph
  {\bibinfo {booktitle} {{Artificial Intelligence in Drug Discovery}}}}\
  (\bibinfo  {publisher} {The Royal Society of Chemistry},\ \bibinfo {year}
  {2020})\BibitemShut {NoStop}%
\bibitem [{\citenamefont {Huo}\ and\ \citenamefont
  {Rupp}(2022)}]{huo_unified_2018}%
  \BibitemOpen
  \bibfield  {author} {\bibinfo {author} {\bibfnamefont {H.}~\bibnamefont
  {Huo}}\ and\ \bibinfo {author} {\bibfnamefont {M.}~\bibnamefont {Rupp}},\
  }\href@noop {} {\bibfield  {journal} {\bibinfo  {journal} {Mach. Learn.: Sci.
  Technol.}\ }\textbf {\bibinfo {volume} {3}},\ \bibinfo {pages} {045017}
  (\bibinfo {year} {2022})}\BibitemShut {NoStop}%
\bibitem [{\citenamefont {Rahaman}\ and\ \citenamefont
  {Gagliardi}(2020)}]{rahaman_2020}%
  \BibitemOpen
  \bibfield  {author} {\bibinfo {author} {\bibfnamefont {O.}~\bibnamefont
  {Rahaman}}\ and\ \bibinfo {author} {\bibfnamefont {A.}~\bibnamefont
  {Gagliardi}},\ }\href@noop {} {\bibfield  {journal} {\bibinfo  {journal} {J.
  Chem. Inf. Model.}\ }\textbf {\bibinfo {volume} {60}},\ \bibinfo {pages}
  {5971} (\bibinfo {year} {2020})}\BibitemShut {NoStop}%
\bibitem [{\citenamefont {Bahlke}\ \emph {et~al.}(2020)\citenamefont {Bahlke},
  \citenamefont {Mogos}, \citenamefont {Proppe},\ and\ \citenamefont
  {Herrmann}}]{bahlke_2020}%
  \BibitemOpen
  \bibfield  {author} {\bibinfo {author} {\bibfnamefont {M.~P.}\ \bibnamefont
  {Bahlke}}, \bibinfo {author} {\bibfnamefont {N.}~\bibnamefont {Mogos}},
  \bibinfo {author} {\bibfnamefont {J.}~\bibnamefont {Proppe}},\ and\ \bibinfo
  {author} {\bibfnamefont {C.}~\bibnamefont {Herrmann}},\ }\href@noop {}
  {\bibfield  {journal} {\bibinfo  {journal} {J. Phys. Chem. A}\ }\textbf
  {\bibinfo {volume} {124}},\ \bibinfo {pages} {8708} (\bibinfo {year}
  {2020})}\BibitemShut {NoStop}%
\bibitem [{\citenamefont {Lumiaro}\ \emph {et~al.}(2021)\citenamefont
  {Lumiaro}, \citenamefont {Todorovi{\'c}}, \citenamefont {Kurten},
  \citenamefont {Vehkam{\"a}ki},\ and\ \citenamefont {Rinke}}]{lumiaro_2021}%
  \BibitemOpen
  \bibfield  {author} {\bibinfo {author} {\bibfnamefont {E.}~\bibnamefont
  {Lumiaro}}, \bibinfo {author} {\bibfnamefont {M.}~\bibnamefont
  {Todorovi{\'c}}}, \bibinfo {author} {\bibfnamefont {T.}~\bibnamefont
  {Kurten}}, \bibinfo {author} {\bibfnamefont {H.}~\bibnamefont
  {Vehkam{\"a}ki}},\ and\ \bibinfo {author} {\bibfnamefont {P.}~\bibnamefont
  {Rinke}},\ }\href@noop {} {\bibfield  {journal} {\bibinfo  {journal} {Atmos.
  Chem. Phys.}\ }\textbf {\bibinfo {volume} {21}},\ \bibinfo {pages} {13227}
  (\bibinfo {year} {2021})}\BibitemShut {NoStop}%
\bibitem [{\citenamefont {Pedregosa}\ \emph {et~al.}(2011)\citenamefont
  {Pedregosa}, \citenamefont {Varoquaux}, \citenamefont {Gramfort},
  \citenamefont {Michel}, \citenamefont {Thirion}, \citenamefont {Grisel},
  \citenamefont {Blondel}, \citenamefont {Prettenhofer}, \citenamefont {Weiss},
  \citenamefont {Dubourg}, \citenamefont {Vanderplas}, \citenamefont {Passos},
  \citenamefont {Cournapeau}, \citenamefont {Brucher}, \citenamefont {Perrot},\
  and\ \citenamefont {Duchesnay}}]{scikit-learn}%
  \BibitemOpen
  \bibfield  {author} {\bibinfo {author} {\bibfnamefont {F.}~\bibnamefont
  {Pedregosa}}, \bibinfo {author} {\bibfnamefont {G.}~\bibnamefont
  {Varoquaux}}, \bibinfo {author} {\bibfnamefont {A.}~\bibnamefont {Gramfort}},
  \bibinfo {author} {\bibfnamefont {V.}~\bibnamefont {Michel}}, \bibinfo
  {author} {\bibfnamefont {B.}~\bibnamefont {Thirion}}, \bibinfo {author}
  {\bibfnamefont {O.}~\bibnamefont {Grisel}}, \bibinfo {author} {\bibfnamefont
  {M.}~\bibnamefont {Blondel}}, \bibinfo {author} {\bibfnamefont
  {P.}~\bibnamefont {Prettenhofer}}, \bibinfo {author} {\bibfnamefont
  {R.}~\bibnamefont {Weiss}}, \bibinfo {author} {\bibfnamefont
  {V.}~\bibnamefont {Dubourg}}, \bibinfo {author} {\bibfnamefont
  {J.}~\bibnamefont {Vanderplas}}, \bibinfo {author} {\bibfnamefont
  {A.}~\bibnamefont {Passos}}, \bibinfo {author} {\bibfnamefont
  {D.}~\bibnamefont {Cournapeau}}, \bibinfo {author} {\bibfnamefont
  {M.}~\bibnamefont {Brucher}}, \bibinfo {author} {\bibfnamefont
  {M.}~\bibnamefont {Perrot}},\ and\ \bibinfo {author} {\bibfnamefont
  {E.}~\bibnamefont {Duchesnay}},\ }\href@noop {} {\bibfield  {journal}
  {\bibinfo  {journal} {JMLR}\ }\textbf {\bibinfo {volume} {12}},\ \bibinfo
  {pages} {2825} (\bibinfo {year} {2011})}\BibitemShut {NoStop}%
\bibitem [{\citenamefont {Fawcett}(2006)}]{fawcett_introduction_2006}%
  \BibitemOpen
  \bibfield  {author} {\bibinfo {author} {\bibfnamefont {T.}~\bibnamefont
  {Fawcett}},\ }\href@noop {} {\bibfield  {journal} {\bibinfo  {journal}
  {Pattern Recognit. Lett.}\ }\textbf {\bibinfo {volume} {27}},\ \bibinfo
  {pages} {861} (\bibinfo {year} {2006})}\BibitemShut {NoStop}%
\bibitem [{\citenamefont {Citovsky}\ \emph {et~al.}(2021)\citenamefont
  {Citovsky}, \citenamefont {DeSalvo}, \citenamefont {Gentile}, \citenamefont
  {Karydas}, \citenamefont {Rajagopalan}, \citenamefont {Rostamizadeh},\ and\
  \citenamefont {Kumar}}]{citovsky_neurips_2021}%
  \BibitemOpen
  \bibfield  {author} {\bibinfo {author} {\bibfnamefont {G.}~\bibnamefont
  {Citovsky}}, \bibinfo {author} {\bibfnamefont {G.}~\bibnamefont {DeSalvo}},
  \bibinfo {author} {\bibfnamefont {C.}~\bibnamefont {Gentile}}, \bibinfo
  {author} {\bibfnamefont {L.}~\bibnamefont {Karydas}}, \bibinfo {author}
  {\bibfnamefont {A.}~\bibnamefont {Rajagopalan}}, \bibinfo {author}
  {\bibfnamefont {A.}~\bibnamefont {Rostamizadeh}},\ and\ \bibinfo {author}
  {\bibfnamefont {S.}~\bibnamefont {Kumar}},\ }\href@noop {} {\bibfield
  {journal} {\bibinfo  {journal} {NeurIPS}\ }\textbf {\bibinfo {volume} {34}},\
  \bibinfo {pages} {11933} (\bibinfo {year} {2021})}\BibitemShut {NoStop}%
\bibitem [{\citenamefont {Kulichenko}\ \emph {et~al.}(2023)\citenamefont
  {Kulichenko}, \citenamefont {Barros}, \citenamefont {Lubbers}, \citenamefont
  {Li}, \citenamefont {Messerly}, \citenamefont {Tretiak}, \citenamefont
  {Smith},\ and\ \citenamefont {Nebgen}}]{kulichenko_uncertainty-driven_2023}%
  \BibitemOpen
  \bibfield  {author} {\bibinfo {author} {\bibfnamefont {M.}~\bibnamefont
  {Kulichenko}}, \bibinfo {author} {\bibfnamefont {K.}~\bibnamefont {Barros}},
  \bibinfo {author} {\bibfnamefont {N.}~\bibnamefont {Lubbers}}, \bibinfo
  {author} {\bibfnamefont {Y.~W.}\ \bibnamefont {Li}}, \bibinfo {author}
  {\bibfnamefont {R.}~\bibnamefont {Messerly}}, \bibinfo {author}
  {\bibfnamefont {S.}~\bibnamefont {Tretiak}}, \bibinfo {author} {\bibfnamefont
  {J.~S.}\ \bibnamefont {Smith}},\ and\ \bibinfo {author} {\bibfnamefont
  {B.}~\bibnamefont {Nebgen}},\ }\href@noop {} {\bibfield  {journal} {\bibinfo
  {journal} {Nat. Comput. Sci.}\ }\textbf {\bibinfo {volume} {3}},\ \bibinfo
  {pages} {230} (\bibinfo {year} {2023})}\BibitemShut {NoStop}%
\bibitem [{\citenamefont {Zhang}\ and\ \citenamefont
  {Lee}(2019)}]{zhang_bayesian_2019}%
  \BibitemOpen
  \bibfield  {author} {\bibinfo {author} {\bibfnamefont {Y.}~\bibnamefont
  {Zhang}}\ and\ \bibinfo {author} {\bibfnamefont {A.~A.}\ \bibnamefont
  {Lee}},\ }\href@noop {} {\bibfield  {journal} {\bibinfo  {journal} {Chem.
  Sci.}\ }\textbf {\bibinfo {volume} {10}},\ \bibinfo {pages} {8154} (\bibinfo
  {year} {2019})}\BibitemShut {NoStop}%
\bibitem [{\citenamefont {Li}\ \emph {et~al.}(2021)\citenamefont {Li},
  \citenamefont {Rao}, \citenamefont {Hassaine}, \citenamefont {Ramakrishnan},
  \citenamefont {Canoy}, \citenamefont {Salimi-Khorshidi}, \citenamefont
  {Mamouei}, \citenamefont {Lukasiewicz},\ and\ \citenamefont
  {Rahimi}}]{li_deep_2021}%
  \BibitemOpen
  \bibfield  {author} {\bibinfo {author} {\bibfnamefont {Y.}~\bibnamefont
  {Li}}, \bibinfo {author} {\bibfnamefont {S.}~\bibnamefont {Rao}}, \bibinfo
  {author} {\bibfnamefont {A.}~\bibnamefont {Hassaine}}, \bibinfo {author}
  {\bibfnamefont {R.}~\bibnamefont {Ramakrishnan}}, \bibinfo {author}
  {\bibfnamefont {D.}~\bibnamefont {Canoy}}, \bibinfo {author} {\bibfnamefont
  {G.}~\bibnamefont {Salimi-Khorshidi}}, \bibinfo {author} {\bibfnamefont
  {M.}~\bibnamefont {Mamouei}}, \bibinfo {author} {\bibfnamefont
  {T.}~\bibnamefont {Lukasiewicz}},\ and\ \bibinfo {author} {\bibfnamefont
  {K.}~\bibnamefont {Rahimi}},\ }\href@noop {} {\bibfield  {journal} {\bibinfo
  {journal} {Sci. Rep.}\ }\textbf {\bibinfo {volume} {11}},\ \bibinfo {pages}
  {20685} (\bibinfo {year} {2021})}\BibitemShut {NoStop}%
\bibitem [{\citenamefont {Zaverkin}\ \emph {et~al.}(2022)\citenamefont
  {Zaverkin}, \citenamefont {Holzmüller}, \citenamefont {Steinwart},\ and\
  \citenamefont {Kästner}}]{zaverkin_exploring_2022}%
  \BibitemOpen
  \bibfield  {author} {\bibinfo {author} {\bibfnamefont {V.}~\bibnamefont
  {Zaverkin}}, \bibinfo {author} {\bibfnamefont {D.}~\bibnamefont
  {Holzmüller}}, \bibinfo {author} {\bibfnamefont {I.}~\bibnamefont
  {Steinwart}},\ and\ \bibinfo {author} {\bibfnamefont {J.}~\bibnamefont
  {Kästner}},\ }\href@noop {} {\bibfield  {journal} {\bibinfo  {journal}
  {Digital Discov.}\ }\textbf {\bibinfo {volume} {1}},\ \bibinfo {pages} {605}
  (\bibinfo {year} {2022})}\BibitemShut {NoStop}%
\bibitem [{\citenamefont {Zhang}\ \emph {et~al.}(2023)\citenamefont {Zhang},
  \citenamefont {Chen}, \citenamefont {Rondinelli},\ and\ \citenamefont
  {Chen}}]{zhang_et-_2023}%
  \BibitemOpen
  \bibfield  {author} {\bibinfo {author} {\bibfnamefont {H.}~\bibnamefont
  {Zhang}}, \bibinfo {author} {\bibfnamefont {W.~W.}\ \bibnamefont {Chen}},
  \bibinfo {author} {\bibfnamefont {J.~M.}\ \bibnamefont {Rondinelli}},\ and\
  \bibinfo {author} {\bibfnamefont {W.}~\bibnamefont {Chen}},\ }\href
  {https://doi.org/10.1063/5.0138913} {\bibfield  {journal} {\bibinfo
  {journal} {Appl. Phys. Rev.}\ }\textbf {\bibinfo {volume} {10}},\ \bibinfo
  {pages} {021403} (\bibinfo {year} {2023})}\BibitemShut {NoStop}%
\bibitem [{\citenamefont {Filstroff}\ \emph {et~al.}(2021)\citenamefont
  {Filstroff}, \citenamefont {Sundin}, \citenamefont {Mikkola}, \citenamefont
  {Tiulpin}, \citenamefont {Kylmäoja},\ and\ \citenamefont
  {Kaski}}]{filstroff_targeted_2021}%
  \BibitemOpen
  \bibfield  {author} {\bibinfo {author} {\bibfnamefont {L.}~\bibnamefont
  {Filstroff}}, \bibinfo {author} {\bibfnamefont {I.}~\bibnamefont {Sundin}},
  \bibinfo {author} {\bibfnamefont {P.}~\bibnamefont {Mikkola}}, \bibinfo
  {author} {\bibfnamefont {A.}~\bibnamefont {Tiulpin}}, \bibinfo {author}
  {\bibfnamefont {J.}~\bibnamefont {Kylmäoja}},\ and\ \bibinfo {author}
  {\bibfnamefont {S.}~\bibnamefont {Kaski}},\ }\href@noop {} {\bibfield
  {journal} {\bibinfo  {journal} {{arXiv}}\ ,\ \bibinfo {pages} {2106.04193}}
  (\bibinfo {year} {2021})}\BibitemShut {NoStop}%
\bibitem [{\citenamefont {Sundin}\ \emph {et~al.}(2019)\citenamefont {Sundin},
  \citenamefont {Schulam}, \citenamefont {Siivola}, \citenamefont {Vehtari},
  \citenamefont {Saria},\ and\ \citenamefont {Kaski}}]{sundin_active_2019}%
  \BibitemOpen
  \bibfield  {author} {\bibinfo {author} {\bibfnamefont {I.}~\bibnamefont
  {Sundin}}, \bibinfo {author} {\bibfnamefont {P.}~\bibnamefont {Schulam}},
  \bibinfo {author} {\bibfnamefont {E.}~\bibnamefont {Siivola}}, \bibinfo
  {author} {\bibfnamefont {A.}~\bibnamefont {Vehtari}}, \bibinfo {author}
  {\bibfnamefont {S.}~\bibnamefont {Saria}},\ and\ \bibinfo {author}
  {\bibfnamefont {S.}~\bibnamefont {Kaski}},\ }\href@noop {} {\bibfield
  {journal} {\bibinfo  {journal} {ICML}\ ,\ \bibinfo {pages} {6046}} (\bibinfo
  {year} {2019})}\BibitemShut {NoStop}%
\bibitem [{\citenamefont {Ghosh}(2020{\natexlab{a}})}]{ghosh_2020_3967308}%
  \BibitemOpen
  \bibfield  {author} {\bibinfo {author} {\bibfnamefont {K.}~\bibnamefont
  {Ghosh}},\ }\href@noop {} {\bibinfo {title} {{MBTR AA}}} (\bibinfo {year}
  {2020}{\natexlab{a}}),\ \bibinfo {note}
  {https://doi.org/10.5281/zenodo.3967308}\BibitemShut {NoStop}%
\bibitem [{\citenamefont {Ghosh}(2020{\natexlab{b}})}]{ghosh_2020_4035923}%
  \BibitemOpen
  \bibfield  {author} {\bibinfo {author} {\bibfnamefont {K.}~\bibnamefont
  {Ghosh}},\ }\href@noop {} {\bibinfo {title} {{MBTR OE62}}} (\bibinfo {year}
  {2020}{\natexlab{b}}),\ \bibinfo {note}
  {https://doi.org/10.5281/zenodo.4035923}\BibitemShut {NoStop}%
\bibitem [{\citenamefont {Ghosh}(2020{\natexlab{c}})}]{ghosh_2020_4035918}%
  \BibitemOpen
  \bibfield  {author} {\bibinfo {author} {\bibfnamefont {K.}~\bibnamefont
  {Ghosh}},\ }\href@noop {} {\bibinfo {title} {{MBTR QM9}}} (\bibinfo {year}
  {2020}{\natexlab{c}}),\ \bibinfo {note}
  {https://doi.org/10.5281/zenodo.4035918}\BibitemShut {NoStop}%
\bibitem [{Note1()}]{Note1}%
  \BibitemOpen
  \bibinfo {note} {\protect \url
  {https://github.com/kunalghosh/Multi_Fidelity_Prediction_GP/tree/testing_runs}}\BibitemShut
  {NoStop}%
\end{thebibliography}%

\end{document}